\documentclass[12pt]{article}
\usepackage[margin=0.9in]{geometry}
\usepackage{amsthm,amsmath,amssymb,amscd,enumerate, graphicx,subcaption,pstricks,bm,booktabs}
\usepackage{setspace,array}
\usepackage{thmtools}
\usepackage{thm-restate}
\usepackage{color}
\usepackage{natbib}
\usepackage{multirow}
\usepackage[mathscr]{euscript}
\usepackage[shortlabels]{enumitem}
\usepackage[colorlinks=true, urlcolor=blue, linkcolor=blue, citecolor=blue]{hyperref}
\usepackage{algpseudocode}
\usepackage{algorithm}
\algrenewcommand\algorithmicindent{0.4em}

\usepackage{natbib}
\usepackage[T1]{fontenc}
\usepackage[utf8]{inputenc}
\usepackage{authblk}
\usepackage{siunitx}
\usepackage{longtable}

\title{Practical considerations for variable screening in the super learner}
\author[1,2,3,*]{Brian~D. Williamson}
\author[4]{Drew King}
\author[2,3]{Ying Huang}
\affil[1]{Biostatistics Division, Kaiser Permanente Washington Health Research Institute}
\affil[2]{Vaccine and Infectious Disease Division, Fred Hutchinson Cancer Research Center}
\affil[3]{Department of Biostatistics, University of Washington}
\affil[4]{Seattle Central College}
\affil[*]{brian.d.williamson@kp.org}

\begin{document}

\maketitle

\begin{abstract}

    Estimating a prediction function is a fundamental component of many data analyses. The super learner ensemble, a particular implementation of stacking, has desirable theoretical properties and has been used successfully in many applications. Dimension reduction can be accomplished by using variable screening algorithms (screeners), including the lasso, within the ensemble prior to fitting other prediction algorithms. However, the performance of a super learner using the lasso for dimension reduction has not been fully explored in cases where the lasso is known to perform poorly. We provide empirical results that suggest that a diverse set of candidate screeners should be used to protect against poor performance of any one screener, similar to the guidance for choosing a library of prediction algorithms for the super learner. These results are further illustrated through the analysis of HIV-1 antibody data.

  \begin{center}{\small \textbf{Keywords:} super learner; ensemble machine learning; variable screening; prediction.}\end{center}
\end{abstract}

\doublespacing

\section{Introduction}\label{sec:intro}

Estimating a prediction function is a fundamental component of statistical data analysis. Based on measured outcome $Y$ and covariates $X$, the goal is to estimate the conditional expectation $E(Y \mid X)$. There are many approaches to estimating this regression function, ranging from simple and fully parametric \citep[e.g., generalized linear models;][]{nelder1972} to flexible machine learning approaches, including random forests \citep{breiman2001}, gradient boosted trees \citep{friedman2001}, the lasso \citep{tibshirani1996}, and neural networks \citep{barron1989}. While a single estimator (also referred to as a learner) may be chosen, it can be advantageous to instead consider an ensemble of multiple candidate learners; a large ensemble of flexible learners increases the chance that one learner can approximate the underlying conditional expectation well. 

The super learner (SL) \citep{vanderlaan2007,polley2010} is one such ensemble, and is related to stacking \citep{wolpert1992}. The super learner has been shown to have the same expected loss for predicting the outcome as the oracle estimator, asymptotically \citep{vanderlaan2007}. If both simple and complex algorithms are included in the library of candidate learners, the cross-validation used within the super learner to select the optimal combination of candidate learners to minimize a cross-validated loss function can minimize the risk of overfitting \citep{balzer2021}. The super learner has been used successfully in many applications \citep[see, e.g.,][]{vanderlaan2011,pirracchio2015,petersen2015,magaret2019,carrell2023} and is implemented in several software packages for the R programming language \citep{slpkg,sl3pkg}.

In some settings, it may be of interest to perform variable selection as part of certain candidate learners within the super learner. This includes high-dimensional settings where prediction performance may be improved by reducing the dimension prior to prediction and settings where having a parsimonious set of variables is a goal of the analysis. While recent work has developed general guidelines for specifying a super learner \citep{phillips2023}, the choice of \textit{screening algorithms} (often referred to as \textit{screeners}) has been relatively unexplored. In particular, there are cases where theory suggests that the lasso does not consistently select the most relevant variables \citep{leng2006}. In this article, we explore the use of the lasso as a screener within a super learner ensemble, with the goal of determining if there are cases where the performance of the ensemble is sensitive to possible poor performance of the lasso screener.

\section{Overview of variable screening in the super learner}

\citet{phillips2023} provide a thorough overview of the super learner algorithm, which we briefly summarize here. The super learner takes as input the following: the dataset $\{(X_i,Y_i)\}_{i=1}^n$; a \textit{library} of candidate learners (e.g., random forests, the lasso, neural networks), possibly including combinations with variable screeners (e.g., the lasso) that reduce the dimension of the covariates prior to prediction; a fixed number of cross-validation folds; and a loss function to minimize using cross-validation. The \textit{ensemble super learner} (hereafter eSL) uses a meta-learner to combine the predictions from the candidate learners \citep{phillips2023}. Below, we will refer to a special case of the eSL, which we call the \textit{cSL}: the convex combination of the candidate learners that minimizes the cross-validated loss. The combination weights are greater than or equal to zero by definition. The discrete super learner (dSL) selects the single candidate learner that minimizes the cross-validated loss.

Including variable screeners in the SL library is motivated by the fact that reducing the number of covariates can improve prediction performance in some cases \citep[see, e.g.,][]{tibshirani1996}, for example, high-dimensional settings. Screeners can be broadly categorized as outcome-blind, such as removing one variable from a pair of highly correlated covariates; or based on the outcome-covariate relationship. Examples of this latter category include removing covariates with univariate outcome-correlation-test p-value larger than a threshold; removing covariates with random forest variable importance measure \citep{breiman2001} rank larger than a threshold; or removing covariates with zero estimated lasso coefficient. 

Strategies based on the outcome-covariate relationship, if pursued, should be combined with other algorithms in the SL library and should be evaluated using cross-validation \citep{phillips2023}. In practice, specifying a screener-learner combination results in a new learner, where first the screener is applied and then the learner is applied on the reduced set of covariates. This becomes one of the learners in the SL library, and like any other learner, can either be chosen as part of the optimal combination or assigned zero weight. For example, suppose that $q$ screeners and $\ell$ learners are considered. Then the candidate library could consist of all $q \times \ell$ screener-learner combinations, or a subset of these combinations chosen by the analyst. Below, we will consider all $q \times \ell$ screener-learner pairs. The ensembling step of the super learner assigns non-negative coefficients to each of the screener-learner combinations to create the ensemble learner. 

\section{Numerical experiments}\label{sec:sims}

\subsection{Data-generating mechanisms}
To demonstrate the performance of the SL procedure using different screeners, we consider several data-generating scenarios. In each scenario, our simulated dataset consists of independent replicates of $(X,Y)$, where $X = (X_1, \ldots, X_p)$ is a covariate vector and $Y$ is the outcome of interest. 

We consider a continuous outcome with $Y \mid (X = x) = f(x) + \epsilon$, where $\epsilon \sim N(0, 1)$ independent of $X$; and a binary outcome with $Pr(Y = 1 \mid X = x) = \Phi\{f(x)\}$, where $\Phi$ denotes the cumulative distribution function of the standard normal distribution (so $Y$ follows a probit model). The outcome regression function $f$ is either linear, with $f(x) = x\beta$, or nonlinear, with 
\begin{align*}
    f(x) =& \ \beta_{1} f_1\{c_1(x_1)\} + \beta_{2} f_2\{c_2(x_2), c_3(x_3)\} + \beta_{3} f_3\{c_3(x_3)\} + \beta_{4} f_4\{c_4(x_4)\} \\
  &\ \  + \beta_{5} f_2\{c_5(x_5), c_1(x_1)\} + \beta_{6} f_3\{c_6(x_6)\}, \\
  f_1(x) =& \ \sin\left(\frac{\pi}{4}x \right), f_2(x, y) = xy, f_3(x) = x, f_4(x) = \cos\left(\frac{\pi}{4}x\right).
\end{align*}
The functions $c_1, \ldots, c_6$ scale each variable to have mean zero and standard deviation one. The vector $\beta$ determines the strength of the relationship between outcome and covariates. We define a weak relationship between the outcome and covariates by setting $\beta = (0, 1, 0, 0, 0, 1, \mathbf{0}_{p-6})$, where $p - 6$ variables do not affect the outcome, and a stronger relationship between the outcome and covariates by setting $\beta = (-3, -1, 1, -1.5, -0.5, 0.5, \mathbf{0}_{p-6})$. The covariates follow a multivariate normal distribution with mean zero and covariance matrix $\Sigma$. In the uncorrelated case, $\Sigma$ is the identity matrix. In the correlated case, the variables in the active set (a subset of the first six variables) have correlation 0.9 (in the case of the strong outcome-covariate relationship) or 0.95 (in the case of the weak relationship) while the remaining variables have correlation 0.3. Based on the strength of relationship between outcome and features, whether it is linear or nonlinear, and whether the features are correlated, the outcome rate in the binary case ranges from approximately 13\% to 80\%.

\subsection{Prediction algorithms}
We compared several main prediction algorithms: the lasso, the cSL without including the lasso in its library of candidate learners [referred to as cSL (-lasso)], the cSL including the lasso (referred to as cSL), and the dSL with and without the lasso in its library of candidate learners (referred to as dSL and dSL (-lasso), respectively). For the super learner approaches, we further considered four possible sets of screeners that were fit prior to any learners: no screeners; a lasso screener only; rank correlation, univariate correlation, random forest, and lasso screeners (referred to as ``All'' screeners); and all possible screeners except the lasso [referred to as ``All (-lasso)'']. Tuning parameters for the screeners depended on the total number of features, except for the lasso screener, which always removed variables with zero regression coefficient based on a tuning parameter selected by 10-fold cross-validation. For $p = 10$, we considered a screener that selected all variables and a univariate correlation screener that removed variables with outcome-correlation-test p-value less than 0.2. For $p > 10$, the rank correlation screeners removed variables outside of the top 10, 25, or 50 ranked correlation-test p-values; the univariate correlation screener removed variables with p-value less than 0.2 or 0.4; and the random forest screener removed variables outside of the top 10 or 25 most-important variables, ranked by the random forest variable importance measure \citep{breiman2001}. 

We finalized our cSL specification following the guidelines specified in \citet{phillips2023}. First, because we were interested in estimating the true continuous prediction function for both continuous and binary outcomes, we estimated the $V$-fold cross-validated least squares loss (for continuous outcomes) or log-likelihood loss (for binary outcomes); we then used the non-negative least squares (NNLS) or non-negative log-likelihood metalearner to obtain the optimal convex combination of these learners, respectively. We used stratified cross-validation \citep{kohavi1996} in the binary-outcome case. We used nested cross-validation in all cases to estimate the performance of the cSL and all individual screener-learner pairs. Second, we computed the effective sample size $n_\text{eff}$, and based our choice of $V$ on the flowchart in Figure 1 of \citet{phillips2023}; the values of $V$ are provided below. Finally, our library of screener-learner pairs specified above was designed to be computationally feasible and adapt to high dimensions and different underlying true regression functions.

\begin{table}
    \centering
    \caption{All possible candidate learners for super learners used in the simulations, along with their R implementation, tuning parameter values, and description of the tuning parameters. All tuning parameters besides those listed here are set to their default values. In particular, the random forests are grown with a minimum node size of 5 for continuous outcomes and 1 for binary outcomes and a subsampling fraction of 1; the boosted trees are grown with shrinkage rate of 0.1, and a minimum of 10 observations per node. \\
    ${}^{\dagger}$: $p$ denotes the total number of predictors. }
    \label{tab:sl-algs}
    \begin{tabular}{p{1.25in}|cll}
       Candidate & R & Tuning Parameter & Tuning parameter  \\
       Learner & Implementation & and possible values & description\\ \hline
        Generalized linear models & \texttt{base} & --- & --- \\ \hline
        Random forests & \texttt{ranger} & \texttt{num.trees} = 1000 & Number of trees \\
        & \citep{rangerpkg} & \texttt{min.node.size} & Minimum node size\\
        & & $\in \{5, 20, 50, 100, 250\}$ & \\ 
        \hline
        Gradient boosted & \texttt{xgboost} & \texttt{max.depth} $ = 4$ &  Maximum tree depth\\
        trees & \citep{xgboostpkg} & \texttt{ntree} $\in \{100, 500, 1000\}$ & Number of iterations \\ 
         & & \texttt{shrinkage} $\in \{0.01, 0.1\}$ & Shrinkage \\ \hline
        Multivariate & \texttt{earth} & \texttt{nk} $= \min\{$ & Maximum number of \\
        adaptive & \citep{earthpkg} & $\max\{21, 2p + 1\}$ & model terms before \\ 
        regression splines & & $, 1000\}^\dagger$ & pruning\\\hline
        Lasso & \texttt{glmnet} & $\lambda$, chosen via & $\ell_1$ regularization  \\
        & \citep{glmnetpkg} & 10-fold cross-validation & parameter \\ \hline
    \end{tabular}
\end{table}

\subsection{Experimental overview}
For each $n \in \{200, 500, 1000, 2000, 3000\}$, $p \in \{10,  500\}$, and simulation scenario described above, we generated 1000 random datasets according to this data generating mechanism. For continuous outcomes, $n_\text{eff} = n$; thus, we set $V = 20$ for $n \leq 500$ and set $V = 10$ otherwise. For binary outcomes, $n_\text{eff}$ ranged from 10 (the 5\% incidence outcome at $n = 200$) to 1367 (a 54\% incidence outcome at $n = 3000$). We set $V = n_\text{eff}$ in three cases, and $V = 20$ or $V = 10$ otherwise, depending on the value of $n_\text{eff}$. The exact values of $n_\text{eff}$ and $V$ used are provided in the Supporting Information. We additionally generated a test dataset with sample size 1 million in each replication to estimate the true prediction performance of each prediction function estimated using $V$-fold cross-validation. We measured prediction performance for each algorithm described above using R-squared for continuous outcomes and area under the receiver operating characteristic curve (AUC) and non-negative log likelihood for binary outcomes. For the continuous outcome, R-squared is equivalent to the cross-validated metric that is being optimized: the mean squared error, which is equal to R-squared up to a scaling factor, the outcome variance. For the binary outcome, AUC is often of interest when assessing prediction performance. AUC is not equivalent to non-negative log-likelihood; however, developing a super learner using AUC loss can be unstable in some settings.

\subsection{Results}

We display the results under a strong outcome-feature relationship in Figures~\ref{fig:continuous_strong} and \ref{fig:binary_strong}. Focusing first on a continuous outcome, when the outcome-feature relationship is linear (Figure~\ref{fig:continuous_strong} left column), all estimators have prediction performance converging quickly to the best-possible prediction performance as the sample size increases. In small samples with a linear relationship, removing the lasso from the SL library results in decreased performance. When the outcome-feature relationship is nonlinear (Figure~\ref{fig:continuous_strong} right column), the results depend on the variable screeners and algorithm used. The lasso has poor performance regardless of sample size, particularly in the case with correlated features; this is consistent with theory \citep{leng2006}. Also, particularly for large numbers of features (e.g., when $p = 500$), using the lasso screener alone within a super learner degrades performance, while using a large library of candidate screeners can improve performance over a super learner with no screeners. Having a large library of candidate screeners can protect against poor lasso performance. Results are similar for the binary outcome.

\begin{figure}
    \centering
    \includegraphics[width=1\textwidth]{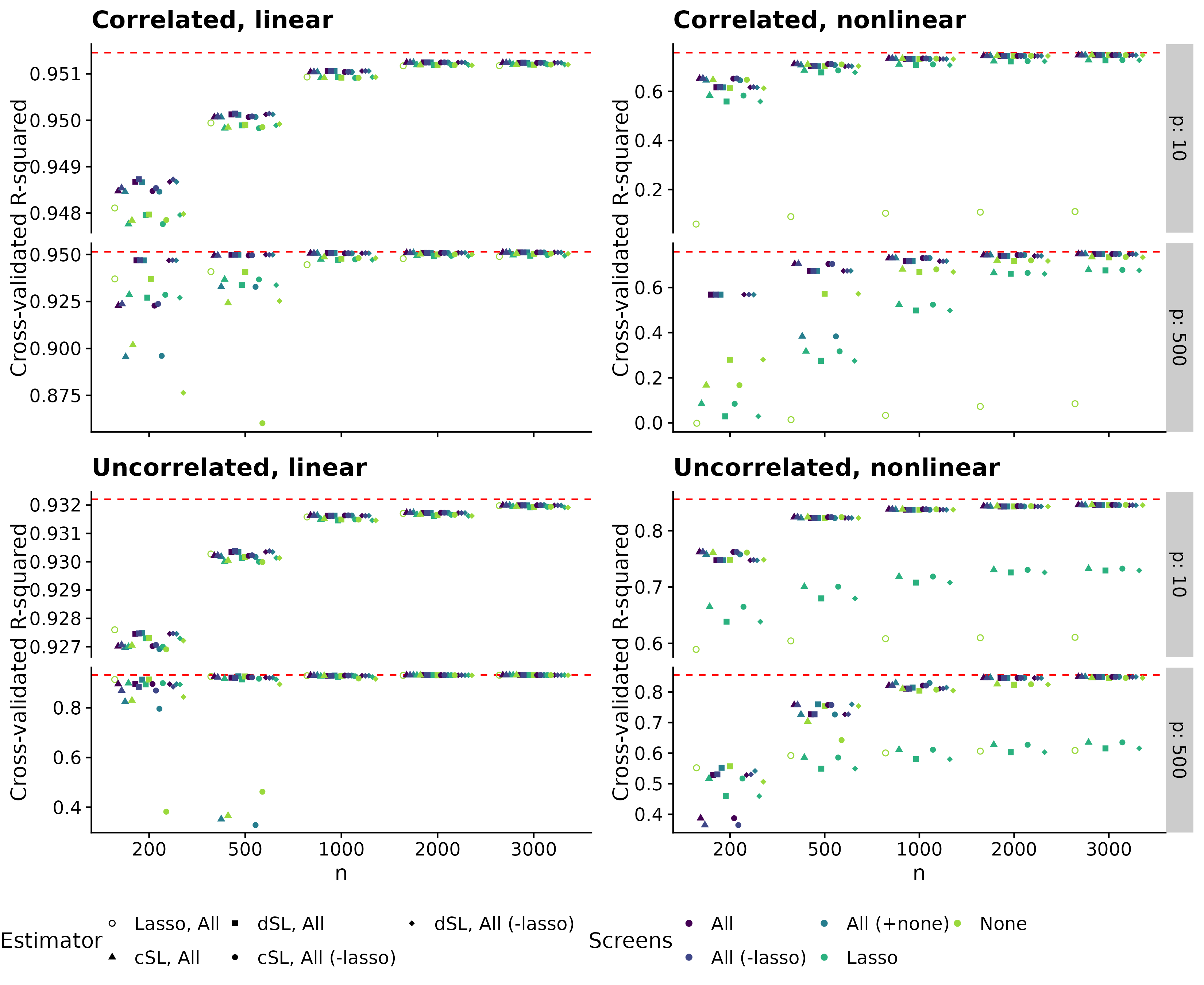}
    \caption{Prediction performance versus sample size $n$, measured using cross-validated R-squared, for predicting a continuous outcome. There is a strong relationship between outcome and features. The top row shows results for correlated features, while the bottom row shows results for uncorrelated features. The left-hand column shows results for a linear outcome-feature relationship, while the right-hand column shows results for a nonlinear outcome-feature relationship. The dashed line denotes the best-possible prediction performance in each setting. Color denotes the variable screeners, while shape denotes the estimator (lasso, convex ensemble super learner [cSL], and discrete super learner [dSL]). Note that the y-axis limits differ between panels.}
    \label{fig:continuous_strong}
\end{figure}

The results under a weak outcome-feature relationship follow similar patterns (Figures~\ref{fig:continuous_weak} and \ref{fig:binary_weak}). In this case, the best-possible prediction performance is lower than in the strong-relationship case, as expected; and a larger sample size is required to achieve prediction performance close to this optimal level.  

\begin{figure}
    \centering
    \includegraphics[width=1\textwidth]{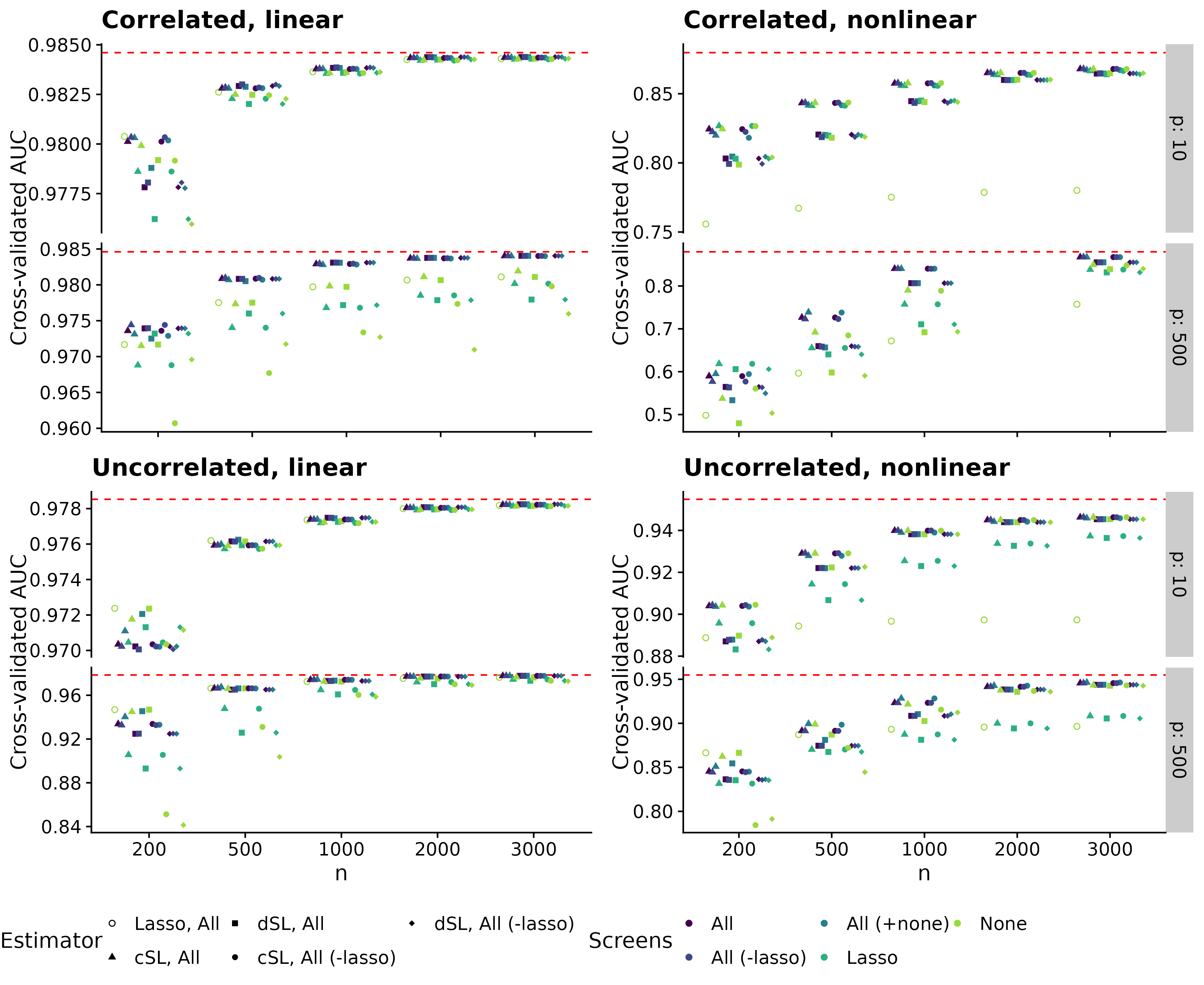}
    \caption{Prediction performance versus sample size $n$, measured using cross-validated AUC, for predicting a binary outcome. There is a strong relationship between outcome and features. The top row shows results for correlated features, while the bottom row shows results for uncorrelated features. The left-hand column shows results for a linear outcome-feature relationship, while the right-hand column shows results for a nonlinear outcome-feature relationship. The dashed line denotes the best-possible prediction performance in each setting. Color denotes the variable screeners, while shape denotes the estimator (lasso, convex ensemble super learner [cSL], and discrete super learner [dSL]). Note that the y-axis limits differ between panels.}
    \label{fig:binary_strong}
\end{figure}

\begin{figure}
    \centering
    \includegraphics[width=1\textwidth]{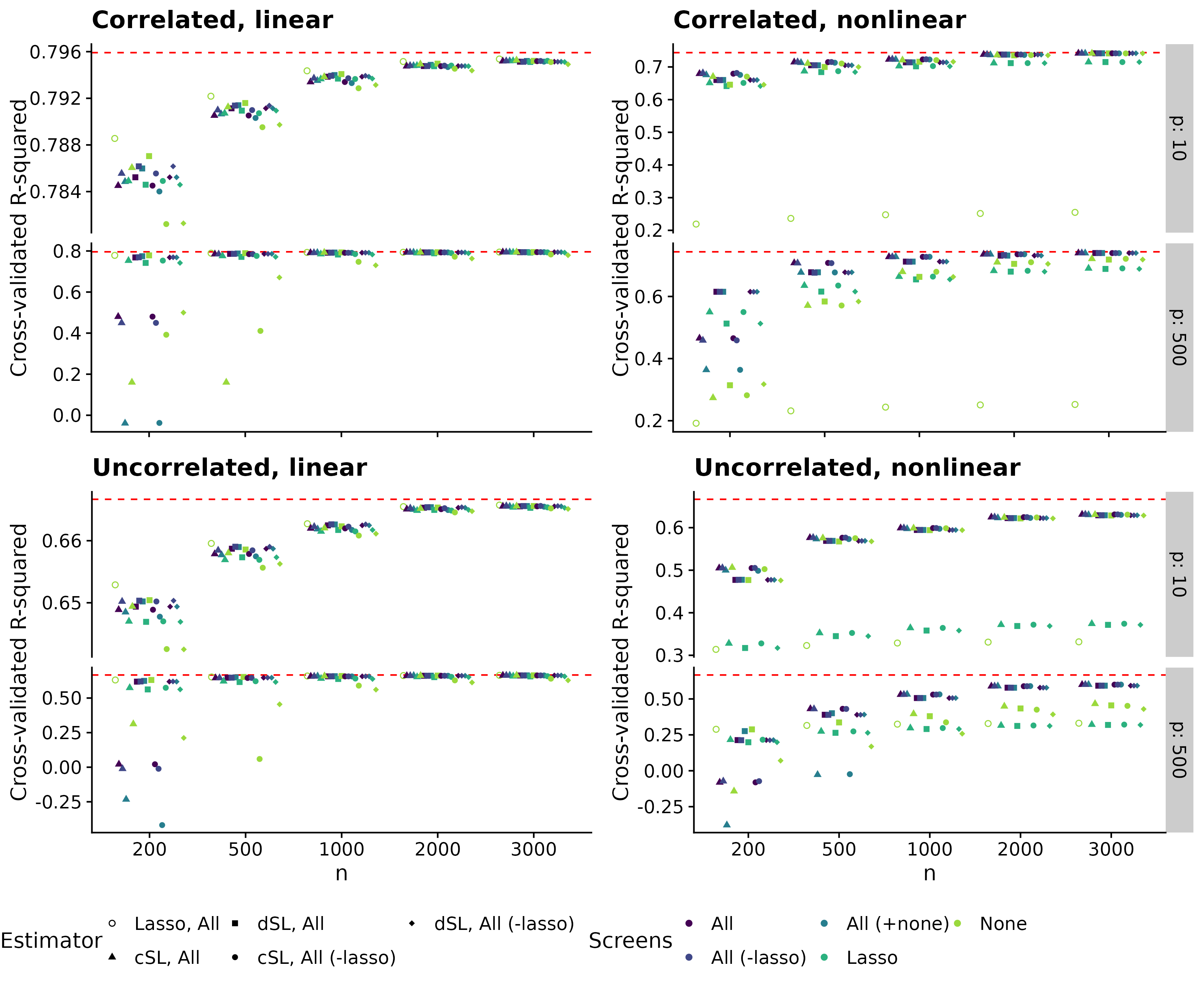}
    \caption{Prediction performance versus sample size $n$, measured using cross-validated R-squared, for predicting a continuous outcome. There is a weak relationship between outcome and features. The top row shows results for correlated features, while the bottom row shows results for uncorrelated features. The left-hand column shows results for a linear outcome-feature relationship, while the right-hand column shows results for a nonlinear outcome-feature relationship. The dashed line denotes the best-possible prediction performance in each setting. Color denotes the variable screeners, while shape denotes the estimator (lasso, convex ensemble super learner [cSL], and discrete super learner [dSL]). Note that the y-axis limits differ between panels.}
    \label{fig:continuous_weak}
\end{figure}

\begin{figure}
    \centering
    \includegraphics[width=1\textwidth]{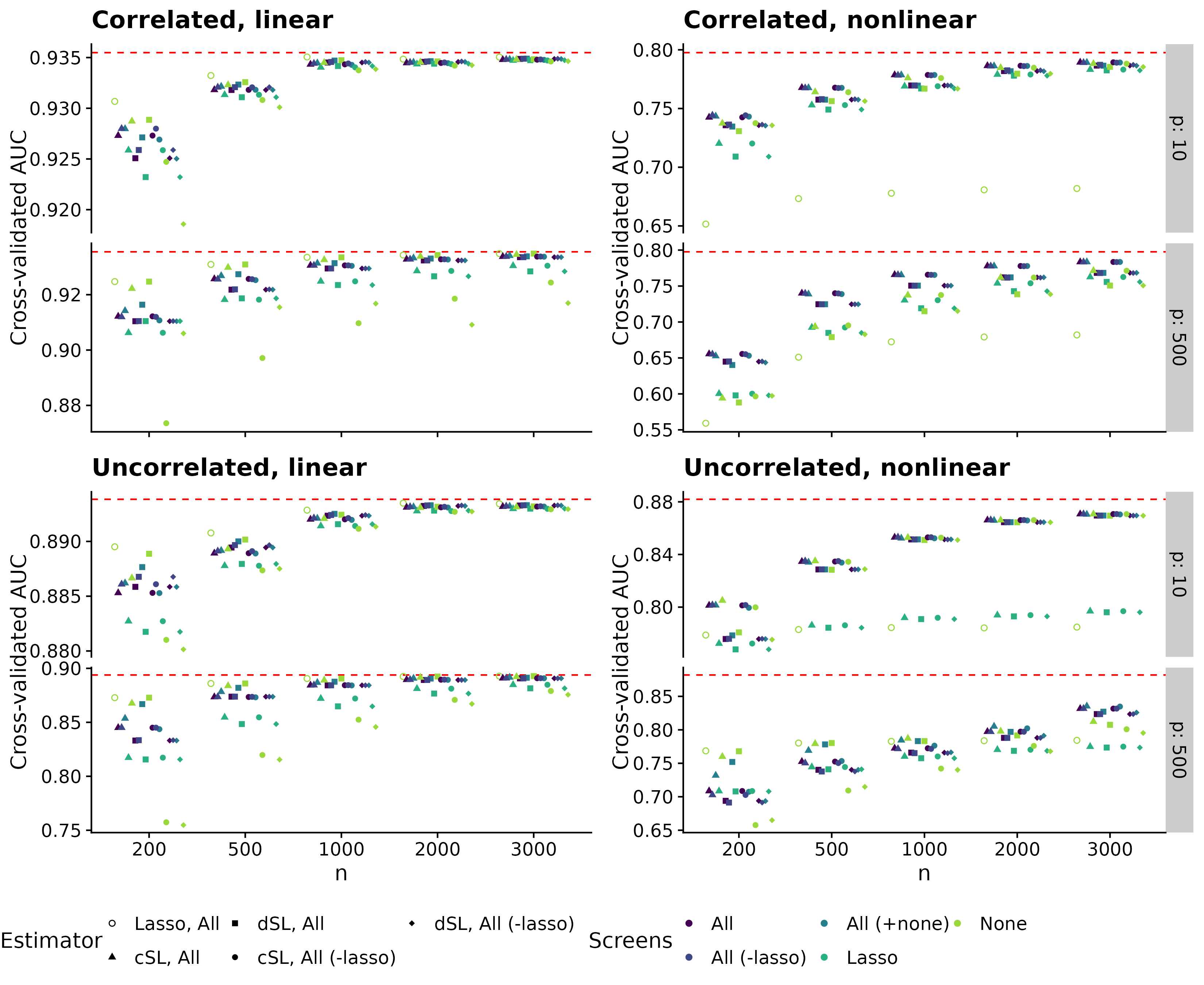}
    \caption{Prediction performance versus sample size $n$, measured using cross-validated AUC, for predicting a binary outcome. There is a weak relationship between outcome and features. The top row shows results for correlated features, while the bottom row shows results for uncorrelated features. The left-hand column shows results for a linear outcome-feature relationship, while the right-hand column shows results for a nonlinear outcome-feature relationship. The dashed line denotes the best-possible prediction performance in each setting. Color denotes the variable screeners, while shape denotes the estimator (lasso, convex ensemble super learner [cSL], and discrete super learner [dSL]). Note that the y-axis limits differ between panels.}
    \label{fig:binary_weak}
\end{figure}

In the Supporting Information, we provide additional results. Results for the binary outcome with respect to non-negative log-likelihood follow similar patterns to those observed here using AUC. We considered further feature dimensions $p$ with a fixed number of cross-validation folds $V$, and found similar results to the primary results presented above. Finally, we present results for $n = 500$ and $p = 2000$ and for candidate learners within the super learner. In the high-dimensional setting, performance follows the same trends across outcomes and estimators as the other $(n, p)$ combinations.

\section{Predicting HIV-1 neutralization susceptibility}\label{sec:data_analysis}

HIV-1 is a genetically diverse pathogen. Broadly neutralizing antibodies (bnAbs) against HIV-1 neutralize a wide array of HIV-1 genetic variants. One such bnAb, VRC01, was recently evaluated in two placebo-controlled randomized trials \citep{corey2021}. Predicting whether or not a given HIV-1 virus is susceptible to neutralization by a bnAb, including VRC01, is an important component of prevention research; several prediction models have been developed recently \citep{hepler2014,buiu2016,hake2017,bricault2019hiv,rawi2019,conti2019,yu2019,magaret2019,williamson2021slapnap,duanuailua2022,williamson2023}.

We analyze HIV-1 envelope (Env) amino acid (AA) sequence data from 611 publicly-available HIV-1 Env pseudoviruses made from blood samples of HIV-1 infected individuals \citep{magaret2019}. In addition to binary indicators of specific AA residues at each position in the Env sequence, the data also include information on the geographic region of origin of the virus, the subtype of the virus, and viral geometry; there are over 800 features in total. We considered two outcomes of interest: the $\log_{10}$-transformed 50\% inhibitory concentration, $\text{IC}_{50}$, defined as the concentration (\si{\micro\gram}/\si{\milli\liter}) of VRC01 necessary to neutralize 50\% of viruses \text{in vitro}, with large values of $\text{IC}_{50}$ indicating resistance to neutralization; and susceptibility to neutralization, defined as the binary indicator that $\text{IC}_{50} < 1$ \si{\micro\gram}/\si{\milli\liter}. For each outcome, we considered the same prediction algorithms and eSL specification as in Section~\ref{sec:sims}. Following \citet{phillips2023}, we set $V = 10$ for both the continuous and binary outcome.

The results are presented in Tables~\ref{tab:data_analysis_main_continuous} and \ref{tab:data_analysis_main_binary}. For both outcomes, some screening tended to be beneficial. Among the analyses that used screeners, using the lasso screener alone resulted in the worst performance for the binary outcome and near the worst for the continuous outcome. Again, for both outcomes, having a large set of screeners protected against poor lasso performance; the lasso performed worse than the cSL or dSL for both outcomes. The lasso had a cross-validated (CV) R-squared for the continuous outcome of 0.331 with a 95\% confidence interval (CI) of [0.305, 0.358], and a CV AUC for the binary outcome of 0.757 [0.633, 0.882]. For the continuous outcome, the largest point estimate of CV R-squared was achieved by the cSL with all screeners, including the lasso; the CV R-squared was 0.394 [0.371, 0.417]. The best-performing dSL was in the case with all screeners but the lasso, with CV R-squared 0.391 [0.372, 0.411].  For the binary outcome, the largest CV AUC for the cSL was 0.826 [0.723, 0.929] in the case with all screeners but the lasso; for the dSL, the largest CV AUC was 0.837 [0.737, 0.936] in the case with no screeners. In the Supporting Information, we present cross-validated performance for the candidate learners in each cSL; cross-validated negative log-likelihood loss for the binary susceptibility outcome; and the cSL coefficients and dSLs for each cross-validation fold.

\begin{table}
\centering
\caption{Estimates of cross-validated R-squared for the continuous $\text{IC}_{50}$ outcome, for the convex ensemble super learner (cSL), the discrete super learner (dSL), and the lasso, under each combination of learners and screeners. For screeners, `None' denotes no screeners; `Lasso' denotes only a lasso screener; `All (-lasso)' denotes random forest, rank-correlation, and correlation-test p-value screening; `All' denotes these three screener types plus the lasso; and `All (+none)' denotes all screeners plus the `none' screener.\label{tab:data_analysis_main_continuous}}
\centering
\resizebox{\ifdim\width>\linewidth\linewidth\else\width\fi}{!}{
\fontsize{9}{11}\selectfont
\begin{tabular}[t]{lllrrl}
\toprule
Learners & Screeners & Algorithm & Min & Max & Point estimate [95\% CI]\\
\midrule
All & None & cSL & 0.208 & 0.501 & 0.373 [0.353, 0.393]\\
All & None & dSL & 0.058 & 0.491 & 0.366 [0.347, 0.385]\\
All & None & lasso & 0.331 & 0.331 & 0.331 [0.305, 0.358]\\
All & Lasso & cSL & 0.175 & 0.527 & 0.388 [0.364, 0.414]\\
All & Lasso & dSL & 0.173 & 0.516 & 0.387 [0.366, 0.409]\\
All & All (-lasso) & cSL & 0.182 & 0.535 & 0.390 [0.370, 0.411]\\
All & All (-lasso) & dSL & 0.192 & 0.519 & 0.391 [0.372, 0.411]\\
All & All & cSL & 0.180 & 0.545 & 0.394 [0.371, 0.417]\\
All & All & dSL & 0.173 & 0.516 & 0.387 [0.365, 0.409]\\
All & All (+none) & cSL & 0.203 & 0.533 & 0.378 [0.354, 0.403]\\
All & All (+none) & dSL & 0.173 & 0.516 & 0.387 [0.365, 0.409]\\
\bottomrule
\end{tabular}}
\end{table}

\begin{table}
\centering
\caption{Estimates of cross-validated AUC for the binary sensitivity outcome, for the convex ensemble super learner (cSL), the discrete super learner (dSL), and the lasso, under each combination of learners and screeners. For screeners, `None' denotes no screeners; `Lasso' denotes only a lasso screener; `All (-lasso)' denotes random forest, rank-correlation, and correlation-test p-value screening; `All' denotes these three screener types plus the lasso; and `All (+none)' denotes all screeners plus the `none' screener.\label{tab:data_analysis_main_binary}}
\centering
\resizebox{\ifdim\width>\linewidth\linewidth\else\width\fi}{!}{
\fontsize{9}{11}\selectfont
\begin{tabular}[t]{lllrrl}
\toprule
Learners & Screeners & Algorithm & Min & Max & Point estimate [95\% CI]\\
\midrule
All & None & cSL & 0.755 & 0.874 & 0.823 [0.719, 0.928]\\
All & None & dSL & 0.763 & 0.895 & 0.837 [0.737, 0.936]\\
All & None & lasso & 0.647 & 0.813 & 0.757 [0.633, 0.882]\\
All & Lasso & cSL & 0.727 & 0.865 & 0.806 [0.696, 0.915]\\
All & Lasso & dSL & 0.730 & 0.897 & 0.811 [0.703, 0.919]\\
All & All (-lasso) & cSL & 0.752 & 0.906 & 0.826 [0.723, 0.929]\\
All & All (-lasso) & dSL & 0.772 & 0.907 & 0.827 [0.724, 0.929]\\
All & All & cSL & 0.750 & 0.873 & 0.823 [0.719, 0.928]\\
All & All & dSL & 0.772 & 0.897 & 0.826 [0.723, 0.929]\\
All & All (+none) & cSL & 0.746 & 0.879 & 0.825 [0.720, 0.929]\\
All & All (+none) & dSL & 0.772 & 0.897 & 0.829 [0.727, 0.931]\\
\bottomrule
\end{tabular}}
\end{table}

\section{Discussion}\label{sec:discussion}

In this manuscript, we explored the effect of using different combinations of variable screeners within the super learner. We found that both the lasso and the ensemble super learner (cSL) using only a lasso screener had poor prediction performance when the outcome-feature relationship was nonlinear; in other words, in the case where the lasso is misspecified. However, if a sufficiently rich set of candidate screeners were included, then including the lasso as a candidate screener did not degrade performance. These results held for both continuous and binary outcomes, and for both strong and weak relationships between the outcome and features. Add something about the dSL. In an analysis of 611 HIV-1 envelope protein pseudoviruses with over 800 features, we found similar results to the simulations. There, the dSL tended to result in performance similar to the cSL. 

Taken together, the results suggest that some caution must be used when specifying screeners within a super learner, but that a sufficiently large set of candidate screeners can protect against misspecification of a given screener. This guidance is similar to the  guidance to specify a diverse set of learners in a super learner \citep{phillips2023}, and can be viewed as complementary, since an algorithm-screener pair defines a new candidate learner.

\section*{Acknowledgements}\label{sec:acknowledgements}
This work was supported by the National Institutes of Health (NIH) grants R01CA277133, R37AI054165, R01GM106177, U24CA086368 and S10OD028685. The opinions expressed in this article are those of the authors and do not necessarily represent the official views of the NIH.

\section*{Supporting Information}
Code to to reproduce all numerical experiments and the data analysis is available on GitHub at \url{github.com/bdwilliamson/sl_screening_supplementary}.

\renewcommand{\thesection}{S\arabic{section}}
\renewcommand{\thetable}{S\arabic{table}}  
\renewcommand{\thefigure}{S\arabic{figure}}
\renewcommand{\figurename}{Supplemental Figure} 
\renewcommand{\tablename}{Supplemental Table} 
\renewcommand{\theequation}{S\arabic{equation}}
\setcounter{table}{0}
\setcounter{figure}{0}
\setcounter{tocdepth}{2}
\setcounter{section}{0}
\setcounter{equation}{0}

\section{Numerical experiments}

\subsection{Values of $n_\text{eff}$ and $V$}

In Table~\ref{tab:neff-v}, we display the effective sample size $n_\text{eff}$ and the number of cross-validation folds $V$ used in both the outer (to estimate cross-validated prediction performance of the convex ensemble super learner (cSL)) and inner (to select the weights in the cSL) cross-validation for each sample size in the binary-outcome settings. In the continuous-outcome settings, since $n_\text{eff} = n$, we set $V = 20$ for $n \leq 500$ and set $V = 10$ for $n > 500$. All values of $V$ were chosen following \citet{phillips2023}.

\begin{table}[h]
    \centering
    {\fontsize{9pt}{7.2}\selectfont
    \begin{tabular}{c|ccccccccc}
          & \multicolumn{9}{c}{($n_\text{eff}$, $V$)} \\
      $n$ & Scenario 1 & Scenario 2 & Scenario 3 & Scenario 4 & Scenario 5 & Scenario 6 & Scenario 7 & Scenario 8 & Scenario 9 \\ \hline
        200 & (89, 20) & (77, 20) & (81, 20) & (76, 20) & (91, 20) & (82, 20) & (27, 27) & (41, 20) & (10, 10)\\
        500 & (224, 20) & (193, 20) & (204, 20) & (190, 20) & (227, 20) & (205, 20) & (67, 20) & (103, 20) & (25, 25)\\
        1000 & (448, 20) & (387, 20) & (409, 20) & (380, 20) & (455, 20) & (410, 20) & (135, 20) & (206, 20) & (51, 20)\\
        2000 & (897, 10) & (774, 10) & (818, 10) & (761, 10) & (911, 10) & (821, 10) & (270, 20) & (412, 20) & (103, 20)\\
        3000 & (1346, 10) & (1161, 10) & (1227, 10) & (1142, 10) & (1367, 10) & (1231, 10) & (406, 20) & (618, 10) & (155, 20)\\
    \end{tabular}
    \caption{Values of $(n_\text{eff}, V)$ for each sample size and data-generating scenario in the binary-outcome simulations.}
    \label{tab:neff-v}
    }
\end{table}

\subsection{Additional numerical results}

We first present results for the binary-outcome simulations in the main manuscript with respect to non-negative log likelihood.

\begin{figure}
    \centering
    \includegraphics[width=1\textwidth]{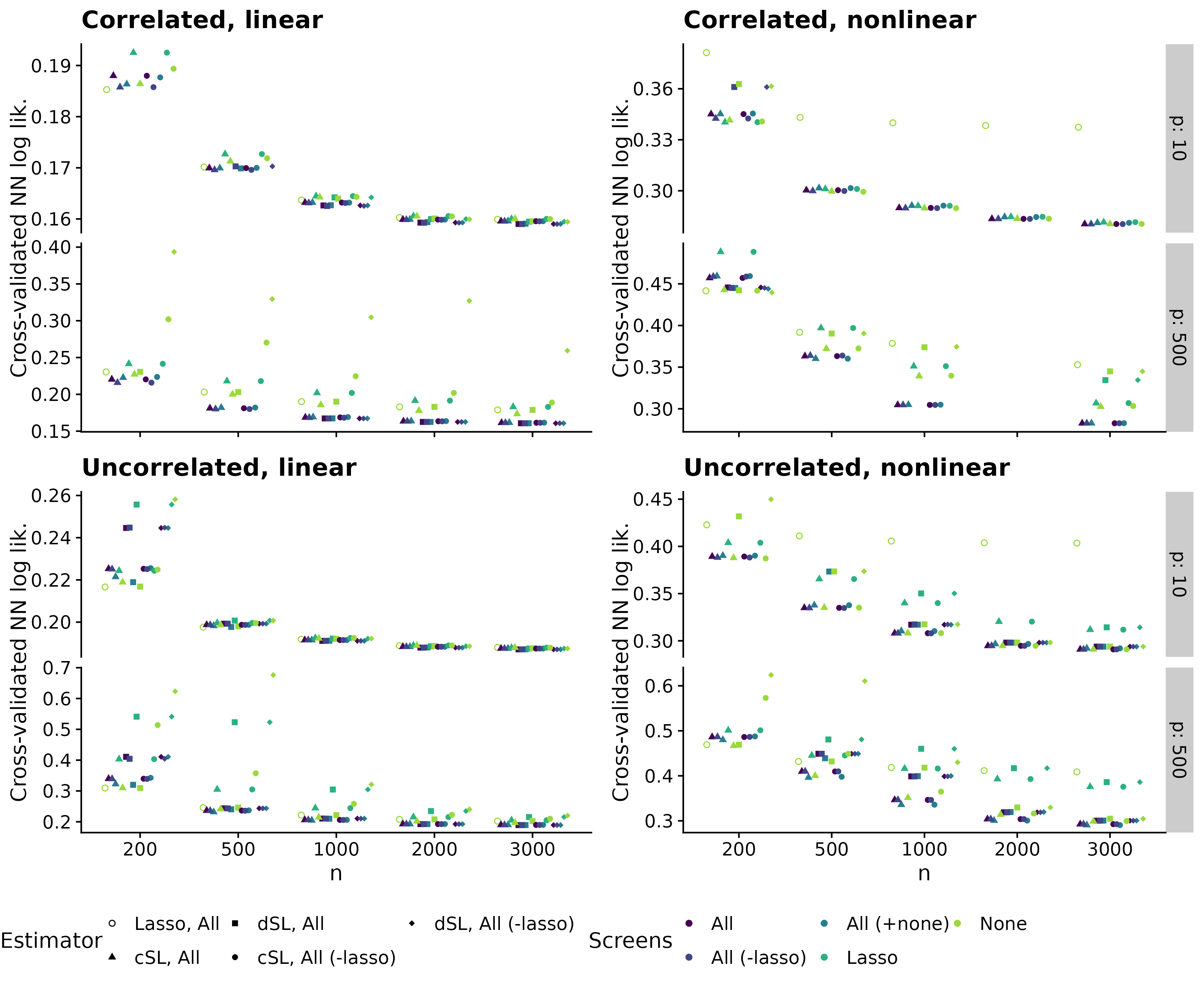}
    \caption{Prediction performance versus sample size $n$, measured using cross-validated non-negative log likelihood (NN log lik.), for predicting a binary outcome. There is a strong relationship between outcome and features. The top row shows results for correlated features, while the bottom row shows results for uncorrelated features. The left-hand column shows results for a linear outcome-feature relationship, while the right-hand column shows results for a nonlinear outcome-feature relationship. The dashed line denotes the best-possible prediction performance in each setting. Color denotes the variable screeners, while shape denotes the estimator (lasso, convex ensemble super learner [cSL], and discrete super learner [dSL]). Note that the y-axis limits differ between panels.}
    \label{fig:binary_strong_nnloglik}
\end{figure}

\begin{figure}
    \centering
    \includegraphics[width=1\textwidth]{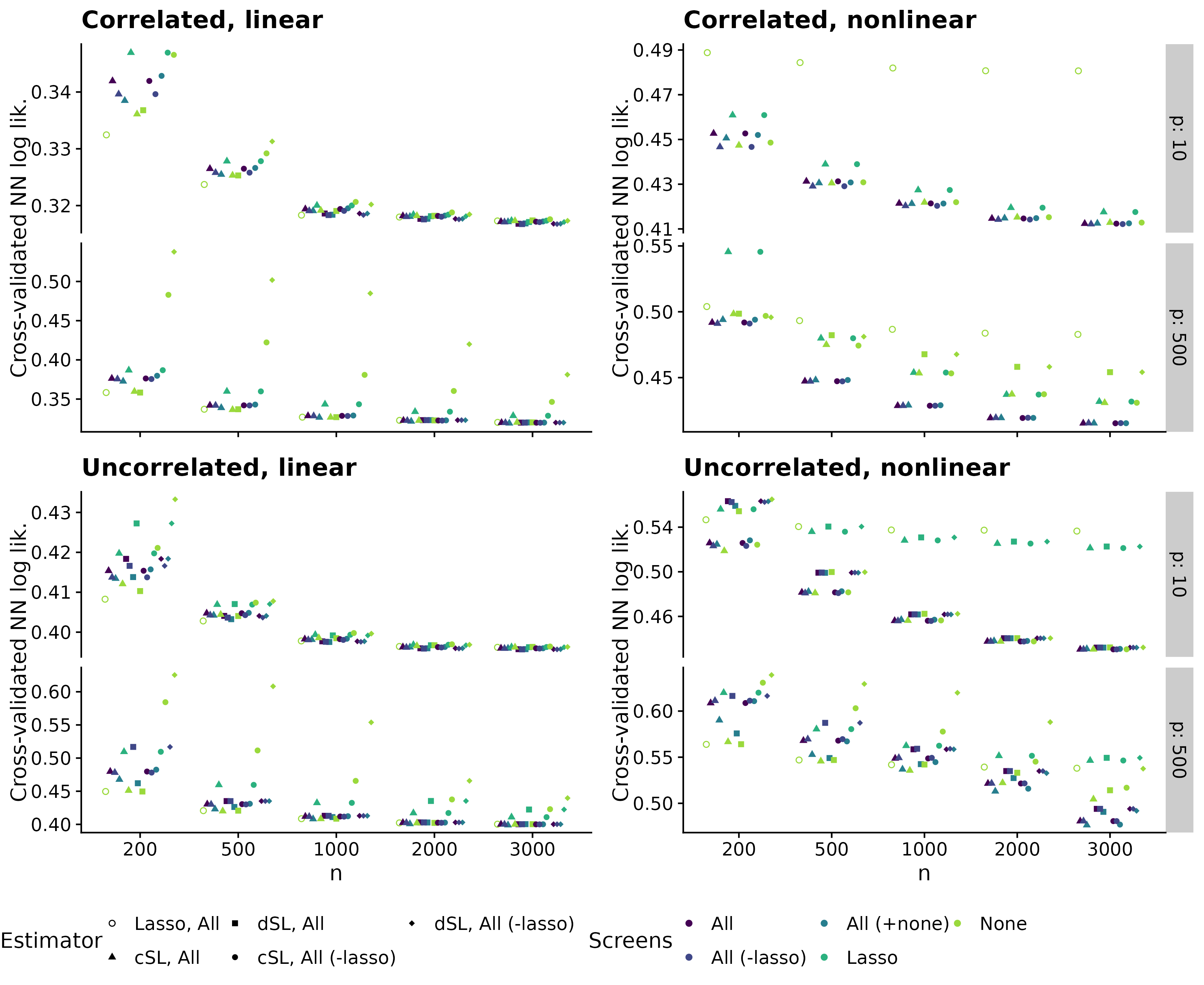}
    \caption{Prediction performance versus sample size $n$, measured using cross-validated non-negative log likelihood (NN log lik.), for predicting a binary outcome. There is a weak relationship between outcome and features. The top row shows results for correlated features, while the bottom row shows results for uncorrelated features. The left-hand column shows results for a linear outcome-feature relationship, while the right-hand column shows results for a nonlinear outcome-feature relationship. The dashed line denotes the best-possible prediction performance in each setting. Color denotes the variable screeners, while shape denotes the estimator (lasso, convex ensemble super learner [cSL], and discrete super learner [dSL]). Note that the y-axis limits differ between panels.}
    \label{fig:binary_weak_nnloglik}
\end{figure}

We provide results for $n = 500$ and $p = 2000$, along with results for each of the candidate learners in the super learner, in Tables~\ref{tab:pred_perf_gaussian_strong_main}--\ref{tab:pred_perf_binomial_rare_strong_matched}. In the high-dimensional case, performance tends to be worse than at smaller $p$ (or with $n > p$), but this is unsurprising. Otherwise, the trends are similar.

\begingroup\fontsize{10}{12}\selectfont

\begin{longtable}[t]{r>{\raggedright\arraybackslash}p{5em}>{\raggedright\arraybackslash}p{5em}>{\raggedright\arraybackslash}p{5em}>{\raggedright\arraybackslash}p{5em}>{\raggedright\arraybackslash}p{5em}>{\raggedright\arraybackslash}p{5em}>{\raggedleft\arraybackslash}p{5em}>{\raggedleft\arraybackslash}p{5em}}
\caption{\label{tab:pred_perf_gaussian_strong_main}Cross-validated R-squared for all candidate learners  in the super learner ensemble  (and the cSL and dSL for p = 2000).  Outcome type: continuous, relationship: strong.}\\
\toprule
n & p & Outcome relationship & Feature correlation & Candidate Learners & Candidate Screeners & Learner & Prediction performance & Optimal prediction performance\\
\midrule
\endfirsthead
\caption[]{Estimates of cross-validated prediction performance for all screener-learner pairs within the super learner. \textit{(continued)}}\\
\toprule
n & p & Outcome relationship & Feature correlation & Candidate Learners & Candidate Screeners & Learner & Prediction performance & Optimal prediction performance\\
\midrule
\endhead

\endfoot
\bottomrule
\endlastfoot
500 & 2000 & Nonlinear & Uncorrelated & All & All & dSL & 0.634 & 0.856\\
500 & 2000 & Nonlinear & Uncorrelated & All & All (+none) & dSL & 0.736 & 0.856\\
500 & 2000 & Nonlinear & Uncorrelated & All & All (-lasso) & dSL & 0.634 & 0.856\\
500 & 2000 & Nonlinear & Uncorrelated & All & Lasso & dSL & 0.513 & 0.856\\
500 & 2000 & Nonlinear & Uncorrelated & All & None & dSL & 0.737 & 0.856\\
500 & 2000 & Nonlinear & Uncorrelated & All (-lasso) & All & dSL & 0.634 & 0.856\\
500 & 2000 & Nonlinear & Uncorrelated & All (-lasso) & All (+none) & dSL & 0.736 & 0.856\\
500 & 2000 & Nonlinear & Uncorrelated & All (-lasso) & All (-lasso) & dSL & 0.634 & 0.856\\
\end{longtable}
\endgroup{}

\begingroup\fontsize{10}{12}\selectfont

\begin{longtable}[t]{r>{\raggedright\arraybackslash}p{5em}>{\raggedright\arraybackslash}p{5em}>{\raggedright\arraybackslash}p{5em}>{\raggedright\arraybackslash}p{5em}>{\raggedright\arraybackslash}p{5em}>{\raggedright\arraybackslash}p{5em}>{\raggedleft\arraybackslash}p{5em}>{\raggedleft\arraybackslash}p{5em}}
\caption{\label{tab:pred_perf_gaussian_weak_main}Cross-validated R-squared for all candidate learners  in the super learner ensemble  (and the cSL and dSL for p = 2000).  Outcome type: continuous, relationship: weak.}\\
\toprule
n & p & Outcome relationship & Feature correlation & Candidate Learners & Candidate Screeners & Learner & Prediction performance & Optimal prediction performance\\
\midrule
\endfirsthead
\caption[]{Estimates of cross-validated prediction performance for all screener-learner pairs within the super learner. \textit{(continued)}}\\
\toprule
n & p & Outcome relationship & Feature correlation & Candidate Learners & Candidate Screeners & Learner & Prediction performance & Optimal prediction performance\\
\midrule
\endhead

\endfoot
\bottomrule
\endlastfoot
500 & 2000 & Nonlinear & Uncorrelated & All & All & dSL & 0.300 & 0.667\\
500 & 2000 & Nonlinear & Uncorrelated & All & All (+none) & dSL & 0.331 & 0.667\\
500 & 2000 & Nonlinear & Uncorrelated & All & All (-lasso) & dSL & 0.300 & 0.667\\
500 & 2000 & Nonlinear & Uncorrelated & All & Lasso & dSL & 0.250 & 0.667\\
500 & 2000 & Nonlinear & Uncorrelated & All & None & dSL & 0.324 & 0.667\\
500 & 2000 & Nonlinear & Uncorrelated & All (-lasso) & All & dSL & 0.300 & 0.667\\
500 & 2000 & Nonlinear & Uncorrelated & All (-lasso) & All (+none) & dSL & 0.303 & 0.667\\
500 & 2000 & Nonlinear & Uncorrelated & All (-lasso) & All (-lasso) & dSL & 0.300 & 0.667\\
\end{longtable}
\endgroup{}

\begingroup\fontsize{10}{12}\selectfont

\begin{longtable}[t]{r>{\raggedright\arraybackslash}p{5em}>{\raggedright\arraybackslash}p{5em}>{\raggedright\arraybackslash}p{5em}>{\raggedright\arraybackslash}p{5em}>{\raggedright\arraybackslash}p{5em}>{\raggedright\arraybackslash}p{5em}>{\raggedleft\arraybackslash}p{5em}>{\raggedleft\arraybackslash}p{5em}}
\caption{\label{tab:pred_perf_binomial_strong_main}Cross-validated AUC for all candidate learners  in the super learner ensemble  (and the cSL and dSL for p = 2000).  Outcome type: binary, relationship: strong.}\\
\toprule
n & p & Outcome relationship & Feature correlation & Candidate Learners & Candidate Screeners & Learner & Prediction performance & Optimal prediction performance\\
\midrule
\endfirsthead
\caption[]{Estimates of cross-validated prediction performance for all screener-learner pairs within the super learner. \textit{(continued)}}\\
\toprule
n & p & Outcome relationship & Feature correlation & Candidate Learners & Candidate Screeners & Learner & Prediction performance & Optimal prediction performance\\
\midrule
\endhead

\endfoot
\bottomrule
\endlastfoot
500 & 2000 & Nonlinear & Uncorrelated & All & All & dSL & 0.866 & 0.955\\
500 & 2000 & Nonlinear & Uncorrelated & All & All (+none) & dSL & 0.876 & 0.955\\
500 & 2000 & Nonlinear & Uncorrelated & All & All (-lasso) & dSL & 0.866 & 0.955\\
500 & 2000 & Nonlinear & Uncorrelated & All & Lasso & dSL & 0.858 & 0.955\\
500 & 2000 & Nonlinear & Uncorrelated & All & None & dSL & 0.884 & 0.955\\
500 & 2000 & Nonlinear & Uncorrelated & All (-lasso) & All & dSL & 0.866 & 0.955\\
500 & 2000 & Nonlinear & Uncorrelated & All (-lasso) & All (+none) & dSL & 0.866 & 0.955\\
500 & 2000 & Nonlinear & Uncorrelated & All (-lasso) & All (-lasso) & dSL & 0.866 & 0.955\\
\end{longtable}
\endgroup{}

\begingroup\fontsize{10}{12}\selectfont

\begin{longtable}[t]{r>{\raggedright\arraybackslash}p{5em}>{\raggedright\arraybackslash}p{5em}>{\raggedright\arraybackslash}p{5em}>{\raggedright\arraybackslash}p{5em}>{\raggedright\arraybackslash}p{5em}>{\raggedright\arraybackslash}p{5em}>{\raggedleft\arraybackslash}p{5em}>{\raggedleft\arraybackslash}p{5em}}
\caption{\label{tab:pred_perf_binomial_strong_matched}Cross-validated NN log lik. for all candidate learners  in the super learner ensemble  (and the cSL and dSL for p = 2000).  Outcome type: binary, relationship: strong.}\\
\toprule
n & p & Outcome relationship & Feature correlation & Candidate Learners & Candidate Screeners & Learner & Prediction performance & Optimal prediction performance\\
\midrule
\endfirsthead
\caption[]{Estimates of cross-validated prediction performance for all screener-learner pairs within the super learner. \textit{(continued)}}\\
\toprule
n & p & Outcome relationship & Feature correlation & Candidate Learners & Candidate Screeners & Learner & Prediction performance & Optimal prediction performance\\
\midrule
\endhead

\endfoot
\bottomrule
\endlastfoot
500 & 2000 & Nonlinear & Uncorrelated & All & All & dSL & 0.461 & 0.558\\
500 & 2000 & Nonlinear & Uncorrelated & All & All (+none) & dSL & 0.449 & 0.558\\
500 & 2000 & Nonlinear & Uncorrelated & All & All (-lasso) & dSL & 0.461 & 0.558\\
500 & 2000 & Nonlinear & Uncorrelated & All & Lasso & dSL & 0.499 & 0.558\\
500 & 2000 & Nonlinear & Uncorrelated & All & None & dSL & 0.440 & 0.558\\
500 & 2000 & Nonlinear & Uncorrelated & All (-lasso) & All & dSL & 0.461 & 0.558\\
500 & 2000 & Nonlinear & Uncorrelated & All (-lasso) & All (+none) & dSL & 0.461 & 0.558\\
500 & 2000 & Nonlinear & Uncorrelated & All (-lasso) & All (-lasso) & dSL & 0.461 & 0.558\\
\end{longtable}
\endgroup{}

\begingroup\fontsize{10}{12}\selectfont

\begin{longtable}[t]{r>{\raggedright\arraybackslash}p{5em}>{\raggedright\arraybackslash}p{5em}>{\raggedright\arraybackslash}p{5em}>{\raggedright\arraybackslash}p{5em}>{\raggedright\arraybackslash}p{5em}>{\raggedright\arraybackslash}p{5em}>{\raggedleft\arraybackslash}p{5em}>{\raggedleft\arraybackslash}p{5em}}
\caption{\label{tab:pred_perf_binomial_weak_main}Cross-validated AUC for all candidate learners  in the super learner ensemble  (and the cSL and dSL for p = 2000).  Outcome type: binary, relationship: weak.}\\
\toprule
n & p & Outcome relationship & Feature correlation & Candidate Learners & Candidate Screeners & Learner & Prediction performance & Optimal prediction performance\\
\midrule
\endfirsthead
\caption[]{Estimates of cross-validated prediction performance for all screener-learner pairs within the super learner. \textit{(continued)}}\\
\toprule
n & p & Outcome relationship & Feature correlation & Candidate Learners & Candidate Screeners & Learner & Prediction performance & Optimal prediction performance\\
\midrule
\endhead

\endfoot
\bottomrule
\endlastfoot
500 & 2000 & Nonlinear & Uncorrelated & All & All & dSL & 0.729 & 0.882\\
500 & 2000 & Nonlinear & Uncorrelated & All & All (+none) & dSL & 0.777 & 0.882\\
500 & 2000 & Nonlinear & Uncorrelated & All & All (-lasso) & dSL & 0.728 & 0.882\\
500 & 2000 & Nonlinear & Uncorrelated & All & Lasso & dSL & 0.731 & 0.882\\
500 & 2000 & Nonlinear & Uncorrelated & All & None & dSL & 0.779 & 0.882\\
500 & 2000 & Nonlinear & Uncorrelated & All (-lasso) & All & dSL & 0.729 & 0.882\\
500 & 2000 & Nonlinear & Uncorrelated & All (-lasso) & All (+none) & dSL & 0.729 & 0.882\\
500 & 2000 & Nonlinear & Uncorrelated & All (-lasso) & All (-lasso) & dSL & 0.728 & 0.882\\
\end{longtable}
\endgroup{}

\begingroup\fontsize{10}{12}\selectfont

\begin{longtable}[t]{r>{\raggedright\arraybackslash}p{5em}>{\raggedright\arraybackslash}p{5em}>{\raggedright\arraybackslash}p{5em}>{\raggedright\arraybackslash}p{5em}>{\raggedright\arraybackslash}p{5em}>{\raggedright\arraybackslash}p{5em}>{\raggedleft\arraybackslash}p{5em}>{\raggedleft\arraybackslash}p{5em}}
\caption{\label{tab:pred_perf_binomial_weak_matched}Cross-validated NN log lik. for all candidate learners  in the super learner ensemble  (and the cSL and dSL for p = 2000).  Outcome type: binary, relationship: weak.}\\
\toprule
n & p & Outcome relationship & Feature correlation & Candidate Learners & Candidate Screeners & Learner & Prediction performance & Optimal prediction performance\\
\midrule
\endfirsthead
\caption[]{Estimates of cross-validated prediction performance for all screener-learner pairs within the super learner. \textit{(continued)}}\\
\toprule
n & p & Outcome relationship & Feature correlation & Candidate Learners & Candidate Screeners & Learner & Prediction performance & Optimal prediction performance\\
\midrule
\endhead

\endfoot
\bottomrule
\endlastfoot
500 & 2000 & Nonlinear & Uncorrelated & All & None & dSL & 0.550 & 0.335\\
500 & 2000 & Nonlinear & Uncorrelated & All (-lasso) & None & dSL & 0.647 & 0.335\\
500 & 2000 & Nonlinear & Uncorrelated & All & None & Lasso & 0.550 & 0.335\\
500 & 2000 & Nonlinear & Uncorrelated & All & All & cSL & 0.573 & 0.335\\
500 & 2000 & Nonlinear & Uncorrelated & All & All (+none) & cSL & 0.557 & 0.335\\
500 & 2000 & Nonlinear & Uncorrelated & All & All (-lasso) & cSL & 0.575 & 0.335\\
500 & 2000 & Nonlinear & Uncorrelated & All & Lasso & cSL & 0.592 & 0.335\\
500 & 2000 & Nonlinear & Uncorrelated & All & None & cSL & 0.550 & 0.335\\
\end{longtable}
\endgroup{}

\begingroup\fontsize{10}{12}\selectfont

\begin{longtable}[t]{r>{\raggedright\arraybackslash}p{5em}>{\raggedright\arraybackslash}p{5em}>{\raggedright\arraybackslash}p{5em}>{\raggedright\arraybackslash}p{5em}>{\raggedright\arraybackslash}p{5em}>{\raggedright\arraybackslash}p{5em}>{\raggedleft\arraybackslash}p{5em}>{\raggedleft\arraybackslash}p{5em}}
\caption{\label{tab:pred_perf_binomial_rare_strong_main}Cross-validated AUC for all candidate learners  in the super learner ensemble  (and the cSL and dSL for p = 2000).  Outcome type: binary (rare), relationship: strong.}\\
\toprule
n & p & Outcome relationship & Feature correlation & Candidate Learners & Candidate Screeners & Learner & Prediction performance & Optimal prediction performance\\
\midrule
\endfirsthead
\caption[]{Estimates of cross-validated prediction performance for all screener-learner pairs within the super learner. \textit{(continued)}}\\
\toprule
n & p & Outcome relationship & Feature correlation & Candidate Learners & Candidate Screeners & Learner & Prediction performance & Optimal prediction performance\\
\midrule
\endhead

\endfoot
\bottomrule
\endlastfoot
500 & 2000 & Nonlinear & Correlated & All & All & dSL & 0.537 & 0.9\\
500 & 2000 & Nonlinear & Correlated & All & All (+none) & dSL & 0.538 & 0.9\\
500 & 2000 & Nonlinear & Correlated & All & All (-lasso) & dSL & 0.535 & 0.9\\
500 & 2000 & Nonlinear & Correlated & All & Lasso & dSL & 0.561 & 0.9\\
500 & 2000 & Nonlinear & Correlated & All & None & dSL & 0.527 & 0.9\\
500 & 2000 & Nonlinear & Correlated & All (-lasso) & All & dSL & 0.537 & 0.9\\
500 & 2000 & Nonlinear & Correlated & All (-lasso) & All (+none) & dSL & 0.524 & 0.9\\
500 & 2000 & Nonlinear & Correlated & All (-lasso) & All (-lasso) & dSL & 0.535 & 0.9\\
\end{longtable}
\endgroup{}

\begingroup\fontsize{10}{12}\selectfont

\begin{longtable}[t]{r>{\raggedright\arraybackslash}p{5em}>{\raggedright\arraybackslash}p{5em}>{\raggedright\arraybackslash}p{5em}>{\raggedright\arraybackslash}p{5em}>{\raggedright\arraybackslash}p{5em}>{\raggedright\arraybackslash}p{5em}>{\raggedleft\arraybackslash}p{5em}>{\raggedleft\arraybackslash}p{5em}}
\caption{\label{tab:pred_perf_binomial_rare_strong_matched}Cross-validated NN log lik. for all candidate learners  in the super learner ensemble  (and the cSL and dSL for p = 2000).  Outcome type: binary (rare), relationship: strong.}\\
\toprule
n & p & Outcome relationship & Feature correlation & Candidate Learners & Candidate Screeners & Learner & Prediction performance & Optimal prediction performance\\
\midrule
\endfirsthead
\caption[]{Estimates of cross-validated prediction performance for all screener-learner pairs within the super learner. \textit{(continued)}}\\
\toprule
n & p & Outcome relationship & Feature correlation & Candidate Learners & Candidate Screeners & Learner & Prediction performance & Optimal prediction performance\\
\midrule
\endhead

\endfoot
\bottomrule
\endlastfoot
500 & 2000 & Nonlinear & Correlated & All & None & dSL & 0.234 & 0.123\\
500 & 2000 & Nonlinear & Correlated & All (-lasso) & None & dSL & 0.239 & 0.123\\
500 & 2000 & Nonlinear & Correlated & All & None & Lasso & 0.232 & 0.123\\
500 & 2000 & Nonlinear & Correlated & All & All & cSL & 0.242 & 0.123\\
500 & 2000 & Nonlinear & Correlated & All & All (+none) & cSL & 0.241 & 0.123\\
500 & 2000 & Nonlinear & Correlated & All & All (-lasso) & cSL & 0.243 & 0.123\\
500 & 2000 & Nonlinear & Correlated & All & Lasso & cSL & 0.268 & 0.123\\
500 & 2000 & Nonlinear & Correlated & All & None & cSL & 0.233 & 0.123\\
\end{longtable}
\endgroup{}

Next, we provide additional numerical experiments under a larger range of $p$ ($p \in \{10, 50, 100, 200\}$) for all sample sizes considered in the main manuscript, and the first eight data-generating mechanisms (the rare-outcome setting is omitted). We follow a similar approach as in the main manuscript, but used 5-fold nested cross-validation (i.e., 5-fold outer cross-validation to obtain the performance of the cSL and 5-fold inner cross-validation to select the cSL weights) in all cases. Importantly, this is not the recommended $V$ following \citet{phillips2023}; however, this choice made the simulations more computationally feasible.

We display the results under a strong outcome-feature relationship in Figures~\ref{fig:continuous_strong} and \ref{fig:binary_strong}. Focusing first on a continuous outcome, when the outcome-feature relationship is linear (Figure~\ref{fig:continuous_strong} left column), all estimators have prediction performance converging quickly to the best-possible prediction performance as the sample size increases. In small samples with a linear relationship, removing the lasso from the SL library results in decreased performance. When the outcome-feature relationship is nonlinear (Figure~\ref{fig:continuous_strong} right column), the results depend on the variable screeners and algorithm used. The lasso has poor performance regardless of sample size, particularly in the case with correlated features; this is consistent with theory \citep{leng2006}. Also, particularly for large numbers of features (e.g., when $p = 500$), using the lasso screener alone within a super learner degrades performance, while using a large library of candidate screeners can improve performance over a super learner with no screeners. Having a large library of candidate screeners can protect against poor lasso performance. Results are similar for the binary outcome.

\begin{figure}
    \centering
    \includegraphics[width=1\textwidth]{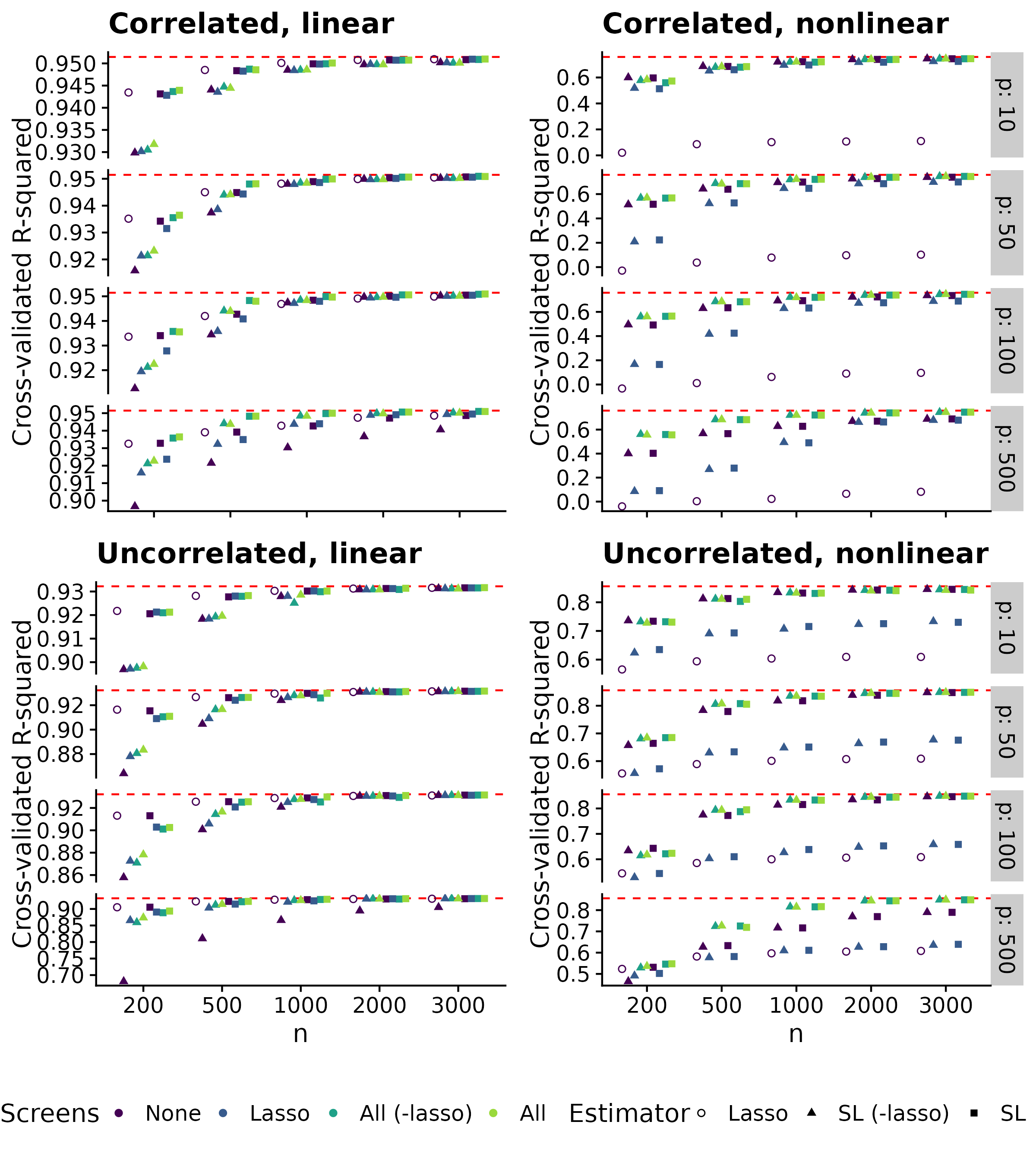}
    \caption{Prediction performance versus sample size $n$, measured using cross-validated R-squared, for predicting a continuous outcome. There is a strong relationship between outcome and features. The top row shows results for correlated features, while the bottom row shows results for uncorrelated features. The left-hand column shows results for a linear outcome-feature relationship, while the right-hand column shows results for a nonlinear outcome-feature relationship. The dashed line denotes the best-possible prediction performance in each setting. Color denotes the variable screeners, while shape denotes the estimator (lasso, discrete super learner (dSL), convex ensemble super learner [cSL], and dSL and cSL without lasso in its library [dSL (-lasso) and cSL (-lasso)]). Note that the y-axis limits differ between panels.}
    \label{fig:continuous_strong}
\end{figure}

The results under a weak outcome-feature relationship follow similar patterns (Figures~\ref{fig:continuous_weak} and \ref{fig:binary_weak}). In this case, the best-possible prediction performance is lower than in the strong-relationship case, as expected; and a larger sample size is required to achieve prediction performance close to this optimal level.  

\begin{figure}
    \centering
    \includegraphics[width=1\textwidth]{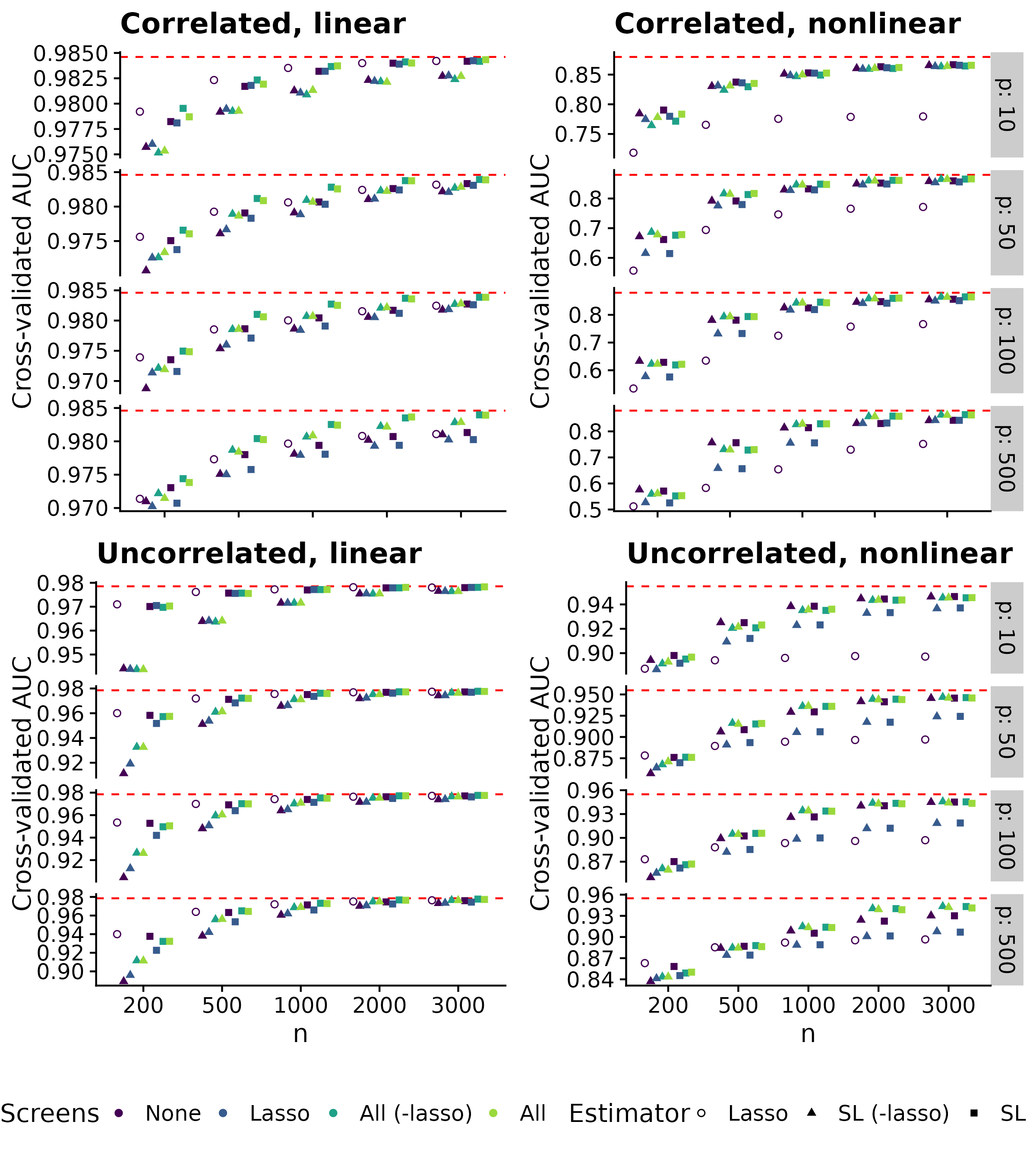}
    \caption{Prediction performance versus sample size $n$, measured using cross-validated AUC, for predicting a binary outcome. There is a strong relationship between outcome and features. The top row shows results for correlated features, while the bottom row shows results for uncorrelated features. The left-hand column shows results for a linear outcome-feature relationship, while the right-hand column shows results for a nonlinear outcome-feature relationship. The dashed line denotes the best-possible prediction performance in each setting. Color denotes the variable screeners, while shape denotes the estimator (lasso, discrete super learner (dSL), convex ensemble super learner [cSL], and dSL and cSL without lasso in its library [dSL (-lasso) and cSL (-lasso)]). Note that the y-axis limits differ between panels.}
    \label{fig:binary_strong}
\end{figure}

\begin{figure}
    \centering
    \includegraphics[width=1\textwidth]{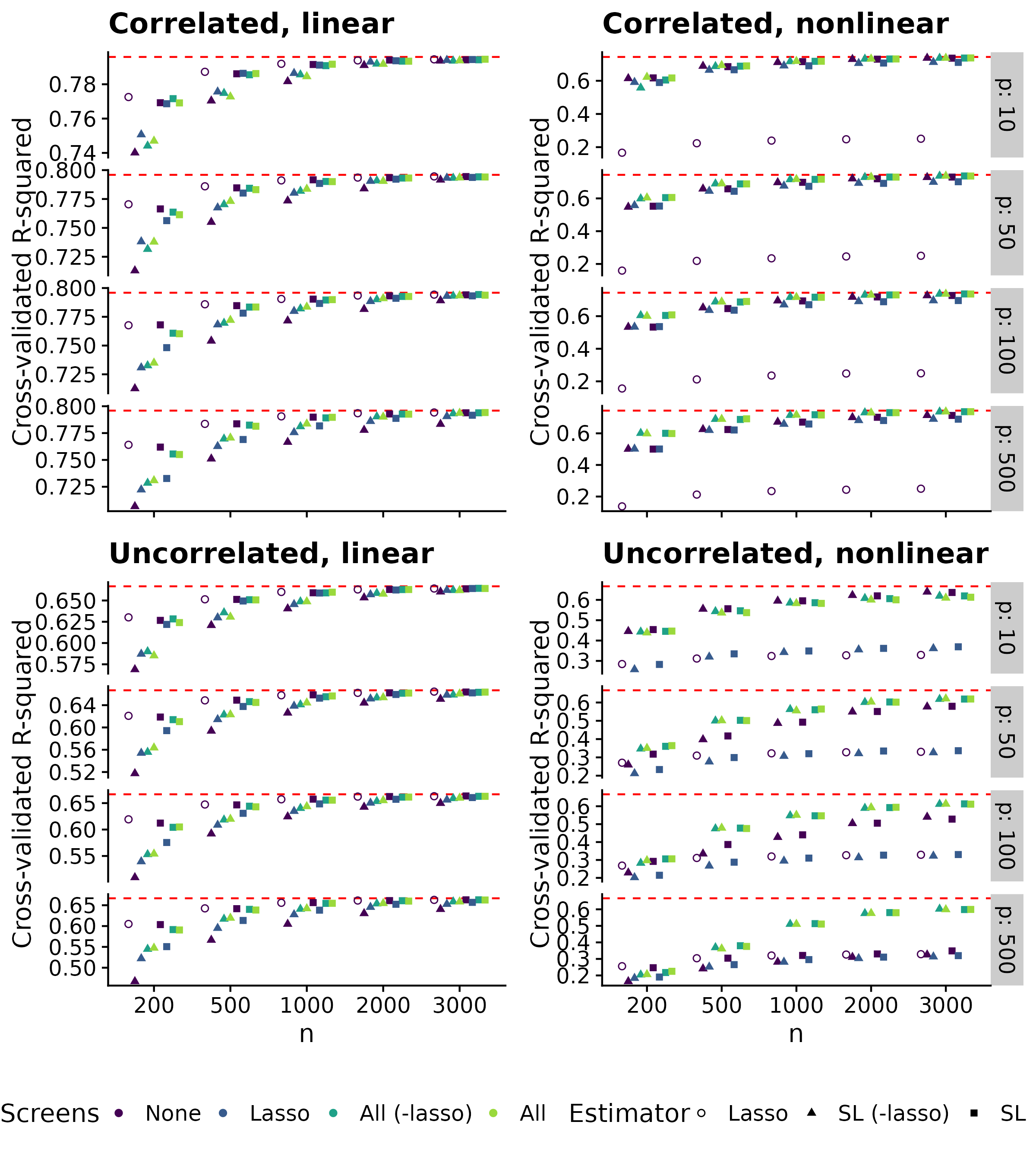}
    \caption{Prediction performance versus sample size $n$, measured using cross-validated R-squared, for predicting a continuous outcome. There is a weak relationship between outcome and features. The top row shows results for correlated features, while the bottom row shows results for uncorrelated features. The left-hand column shows results for a linear outcome-feature relationship, while the right-hand column shows results for a nonlinear outcome-feature relationship. The dashed line denotes the best-possible prediction performance in each setting. Color denotes the variable screeners, while shape denotes the estimator (lasso, discrete super learner (dSL), convex ensemble super learner [cSL], and dSL and cSL without lasso in its library [dSL (-lasso) and cSL (-lasso)]). Note that the y-axis limits differ between panels.}
    \label{fig:continuous_weak}
\end{figure}

\begin{figure}
    \centering
    \includegraphics[width=1\textwidth]{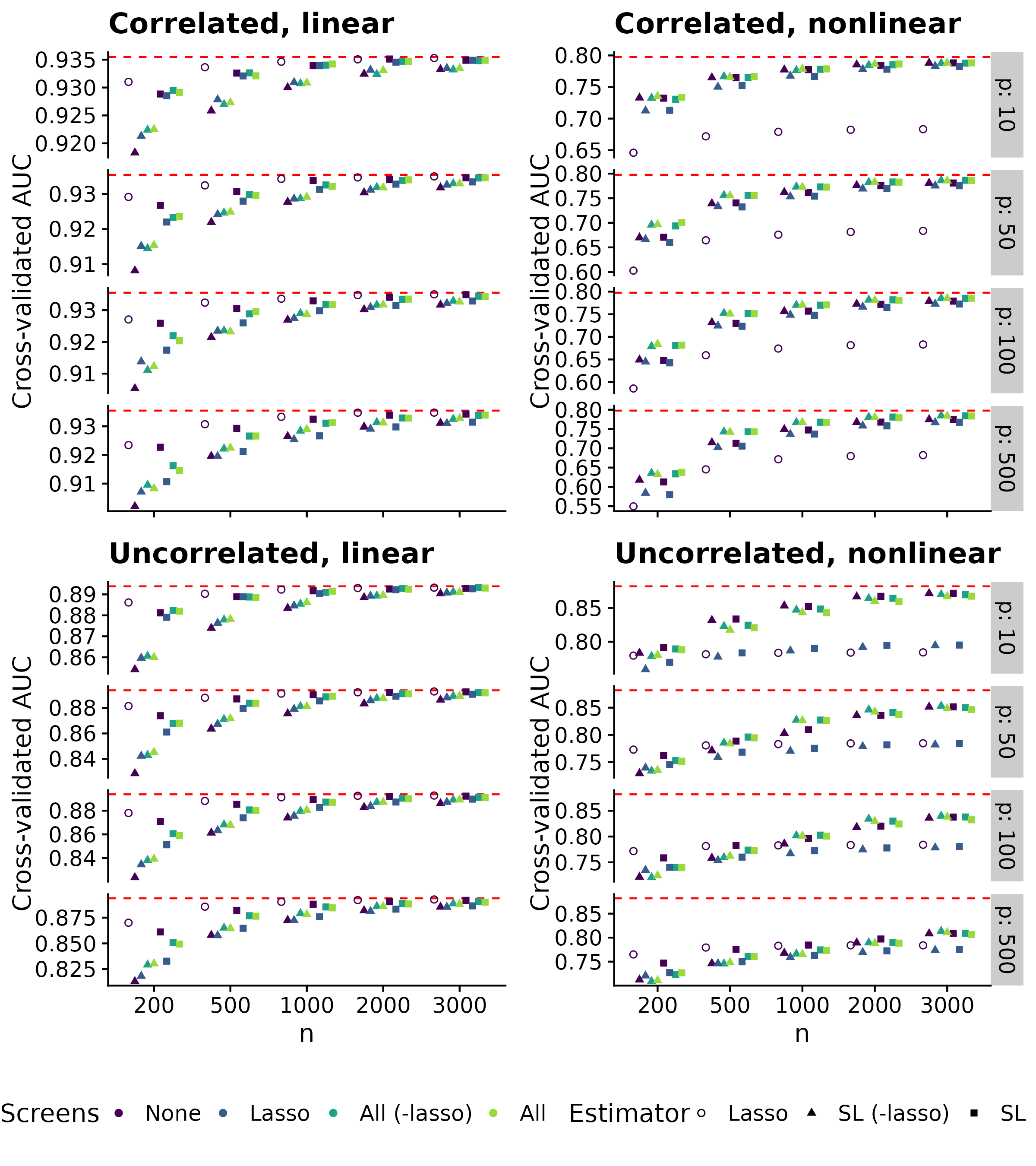}
    \caption{Prediction performance versus sample size $n$, measured using cross-validated AUC, for predicting a binary outcome. There is a weak relationship between outcome and features. The top row shows results for correlated features, while the bottom row shows results for uncorrelated features. The left-hand column shows results for a linear outcome-feature relationship, while the right-hand column shows results for a nonlinear outcome-feature relationship. The dashed line denotes the best-possible prediction performance in each setting. Color denotes the variable screeners, while shape denotes the estimator (lasso, discrete super learner (dSL), convex ensemble super learner [cSL], and dSL and cSL without lasso in its library [dSL (-lasso) and cSL (-lasso)]). Note that the y-axis limits differ between panels.}
    \label{fig:binary_weak}
\end{figure}

\section{Additional results from predicting HIV-1 neutralization susceptibility}

In Tables~\ref{tab:data_analysis_supp_continuous_coefs_dsl} and \ref{tab:data_analysis_supp_binary_coefs_dsl}, we display the super learner coefficients and discrete super learner for each of the 10 cross-validation folds, for both the continuous and binary outcomes.

\begin{table}
\centering
\caption{$\text{IC}_{50}$: Ensemble super learner (cSL) coefficients for each learner-screener pair, for each of the 10 outer cross-validation folds, along with the algorithm selected as the discrete super learner (dSL).\label{tab:data_analysis_supp_continuous_coefs_dsl}}
\centering
\resizebox{\ifdim\width>\linewidth\linewidth\else\width\fi}{!}{
\fontsize{9}{11}\selectfont
\begin{tabular}[t]{>{\raggedright\arraybackslash}p{5em}>{\raggedright\arraybackslash}p{5em}>{\raggedright\arraybackslash}p{5em}>{\raggedright\arraybackslash}p{5em}>{\raggedright\arraybackslash}p{5em}>{\raggedright\arraybackslash}p{5em}>{\raggedright\arraybackslash}p{5em}>{\raggedright\arraybackslash}p{5em}>{\raggedright\arraybackslash}p{5em}>{\raggedright\arraybackslash}p{5em}ll}
\toprule
Learner & Screener & 1 & 2 & 3 & 4 & 5 & 6 & 7 & 8 & 9 & 10\\
\midrule
glm &  & < 0.001 & 0 & < 0.001 & < 0.001 & 0 & < 0.001 & 0 & 0 & 0 & 0\\
glm & corRank.10 & 0 & 0 & 0 & 0 & 0 & 0 & 0 & 0 & 0 & 0\\
glm & corRank.25 & 0 & 0 & 0 & 0 & 0 & 0 & 0 & 0 & 0 & 0\\
glm & corRank.50 & 0 & 0 & 0 & 0 & 0 & 0 & 0 & 0 & 0 & 0.021\\
glm & corP.20 & 0 & 0 & 0 & < 0.001 & 0 & 0 & 0 & 0 & 0.002 & 0\\
glm & corP.40 & 0 & 0 & < 0.001 & 0 & 0.005 & 0 & 0 & 0.002 & 0.003 & < 0.001\\
glm & ranger.10 & 0 & 0 & 0 & 0 & 0 & 0 & 0 & 0 & 0 & 0\\
glm & ranger.25 & 0.015 & 0 & 0 & 0 & 0.026 & 0 & 0.012 & 0 & 0 & < 0.001\\
glm & glmnet & 0.026 & 0.101 & 0.083 & 0.106 & 0.147 & 0.007 & 0.128 & 0.087 & 0 & 0\\
ranger &  & 0.571 & 0.276 & 0 & 0 & 0.572 & 0.191 & 0.486 & 0.305 & 0.287 & 0.407\\
ranger & corRank.10 & 0 & 0 & 0 & 0 & 0 & 0 & 0 & 0 & 0 & 0\\
ranger & corRank.25 & 0 & 0 & 0 & 0 & 0 & 0 & 0 & 0 & 0 & 0\\
ranger & corRank.50 & 0 & 0 & 0 & 0 & 0 & 0 & 0 & 0 & 0 & 0\\
ranger & corP.20 & 0 & 0 & 0.488 & 0 & 0 & 0 & 0.113 & 0 & 0.14 & 0.114\\
ranger & corP.40 & 0.167 & 0 & 0.28 & 0.599 & 0 & 0.402 & 0 & 0.436 & 0 & 0.086\\
ranger & ranger.10 & 0.051 & 0.055 & 0.046 & 0.038 & 0.09 & 0.078 & 0.049 & 0.074 & 0 & 0.019\\
ranger & ranger.25 & 0 & 0 & 0 & 0 & 0.024 & 0 & 0.045 & 0 & 0 & 0\\
ranger & glmnet & 0 & 0.41 & 0 & 0.099 & 0 & 0.23 & 0.048 & 0 & 0.402 & 0.213\\
glmnet &  & 0 & 0 & 0 & 0 & 0 & 0 & 0 & 0 & 0 & 0\\
earth &  & 0 & 0 & 0 & 0.04 & 0.079 & 0.082 & 0 & 0.023 & 0.003 & 0.056\\
earth & corRank.10 & 0 & 0 & 0 & 0 & 0 & 0 & 0 & 0 & 0 & 0.007\\
earth & corRank.25 & 0 & 0 & 0 & 0 & 0.003 & 0 & 0 & 0 & 0 & 0\\
earth & corRank.50 & 0 & 0 & 0 & 0 & 0 & 0.01 & 0 & 0 & 0.028 & 0\\
earth & corP.20 & 0 & 0.015 & 0 & 0.054 & 0 & 0 & 0 & 0 & 0.019 & 0\\
earth & corP.40 & 0.04 & 0.043 & 0.057 & 0.014 & 0 & 0 & 0 & 0 & 0 & 0.015\\
earth & ranger.10 & 0 & 0.046 & 0.044 & 0 & 0.028 & 0 & 0.1 & 0 & 0 & 0.061\\
earth & ranger.25 & 0.097 & 0 & 0 & 0.05 & 0 & 0 & 0.019 & 0.071 & 0.044 & 0\\
earth & glmnet & 0.033 & 0.053 & 0 & 0 & 0.026 & 0 & 0 & 0.001 & 0.072 & 0\\
dSL &  & ranger + glmnet & ranger + glmnet & ranger + corP.20 & ranger + glmnet & ranger + glmnet & ranger + glmnet & ranger + glmnet & ranger + corP.20 & ranger + glmnet & ranger + glmnet\\
\bottomrule
\end{tabular}}
\end{table}

\begin{table}
\centering
\caption{Susceptibility: Ensemble super learner (cSL) coefficients for each learner-screener pair, for each of the 10 outer cross-validation folds, along with the algorithm selected as the discrete super learner (dSL).\label{tab:data_analysis_supp_binary_coefs_dsl}}
\centering
\resizebox{\ifdim\width>\linewidth\linewidth\else\width\fi}{!}{
\fontsize{9}{11}\selectfont
\begin{tabular}[t]{>{\raggedright\arraybackslash}p{5em}>{\raggedright\arraybackslash}p{5em}>{\raggedright\arraybackslash}p{5em}>{\raggedright\arraybackslash}p{5em}>{\raggedright\arraybackslash}p{5em}>{\raggedright\arraybackslash}p{5em}>{\raggedright\arraybackslash}p{5em}>{\raggedright\arraybackslash}p{5em}>{\raggedright\arraybackslash}p{5em}>{\raggedright\arraybackslash}p{5em}ll}
\toprule
Learner & Screener & 1 & 2 & 3 & 4 & 5 & 6 & 7 & 8 & 9 & 10\\
\midrule
glm &  & 0.007 & 0 & 0 & 0 & 0.019 & 0.012 & 0 & 0.005 & 0 & 0\\
glm & corRank.10 & 0.038 & 0.096 & 0.223 & 0.067 & 0.011 & 0.064 & 0.061 & 0 & 0 & 0\\
glm & corRank.25 & 0 & 0 & 0 & 0 & 0 & 0 & 0 & 0.105 & 0 & 0\\
glm & corRank.50 & 0.116 & 0 & 0 & 0.085 & 0 & 0.104 & 0.032 & 0.005 & 0 & 0.065\\
glm & corP.20 & 0.017 & 0.009 & 0.015 & 0.006 & 0 & 0.043 & 0 & < 0.001 & 0 & 0\\
glm & corP.40 & 0.02 & 0 & 0.031 & 0 & 0 & 0.026 & 0.004 & 0 & 0 & 0.018\\
glm & ranger.10 & 0 & 0 & 0 & 0 & 0 & 0 & 0 & 0 & 0 & 0\\
glm & ranger.25 & 0.046 & 0 & 0 & 0.082 & 0.091 & 0 & 0.039 & 0 & 0.034 & 0.044\\
glm & glmnet & 0.012 & 0.04 & 0.02 & 0 & 0.006 & 0 & 0.002 & 0 & 0.036 & 0.011\\
ranger &  & 0.187 & 0.489 & 0.66 & 0.613 & 0 & 0 & 0 & 0 & 0.069 & 0.616\\
ranger & corRank.10 & 0 & 0 & 0 & 0 & 0 & 0 & 0 & 0 & 0 & 0\\
ranger & corRank.25 & 0 & 0 & 0 & 0 & 0 & 0 & 0 & 0 & 0 & 0\\
ranger & corRank.50 & 0 & 0.023 & 0 & 0 & 0 & 0 & 0 & 0 & 0 & 0.189\\
ranger & corP.20 & 0.039 & 0 & 0 & 0 & 0.813 & 0 & 0.653 & 0.497 & 0 & 0\\
ranger & corP.40 & 0.453 & 0 & 0 & 0.095 & 0.003 & 0.706 & 0.093 & 0.312 & 0.726 & 0\\
ranger & ranger.10 & 0 & 0 & 0.009 & 0 & 0 & 0 & 0.028 & 0 & 0 & 0\\
ranger & ranger.25 & 0 & 0 & 0 & 0 & 0 & 0 & 0 & 0 & 0 & 0\\
ranger & glmnet & 0 & 0.264 & 0 & 0 & 0 & 0 & 0 & 0 & 0 & 0\\
glmnet &  & 0 & 0 & 0 & 0 & 0 & 0 & 0 & 0 & 0 & 0\\
earth &  & 0.03 & 0.025 & 0 & 0.023 & 0.025 & 0 & 0.029 & 0.01 & 0.016 & 0\\
earth & corRank.10 & 0.018 & 0 & 0 & 0 & 0 & 0 & 0 & 0.016 & 0.101 & 0\\
earth & corRank.25 & 0 & 0 & 0 & 0 & 0 & 0 & 0 & 0 & 0 & 0\\
earth & corRank.50 & 0 & 0 & 0 & 0 & 0 & 0 & 0 & 0 & 0 & 0.027\\
earth & corP.20 & 0 & 0 & 0.022 & 0.021 & 0 & 0 & 0 & 0 & 0.009 & 0\\
earth & corP.40 & 0.015 & 0.004 & 0.012 & 0.009 & 0.02 & 0.017 & 0.019 & 0.023 & 0.009 & 0.008\\
earth & ranger.10 & 0 & 0.036 & 0 & 0 & 0 & 0 & 0 & 0 & 0 & 0\\
earth & ranger.25 & 0 & 0 & 0 & 0 & 0 & 0 & 0.023 & 0.006 & 0 & 0.022\\
earth & glmnet & 0 & 0.012 & 0.008 & 0 & 0.011 & 0.027 & 0.017 & 0.021 & 0 & 0\\
dSL &  & ranger + corP.40 & ranger + glmnet & ranger + corP.40 & ranger + corP.40 & ranger + corP.20 & ranger + corP.40 & ranger + corP.20 & ranger + corP.20 & ranger + corP.40 & ranger\\
\bottomrule
\end{tabular}}
\end{table}

In Table~\ref{tab:data_analysis_supp_binary}, we display the cross-validated (CV) negative log-likelihood loss for all cSLs, dSLs, and the lasso. The results are largely similar to those for AUC: the lasso has the worst performance (here, largest negative log-likelihood loss), while having a large candidate screener set reduces negative log-likelihood loss.

\begin{table}
\centering
\caption{Estimates of cross-validated negative log likelihood for the binary sensitivity outcome, for the convex ensemble super learner (cSL), the discrete super learner (dSL), and the lasso, under each combination of learners and screeners. For screeners, `None' denotes no screeners; `Lasso' denotes only a lasso screener; `All (-lasso)' denotes random forest, rank-correlation, and correlation-test p-value screening; `All' denotes these three screener types plus the lasso; and `All (+none)' denotes all screeners plus the `none' screener.\label{tab:data_analysis_supp_binary}}
\centering
\resizebox{\ifdim\width>\linewidth\linewidth\else\width\fi}{!}{
\fontsize{9}{11}\selectfont
\begin{tabular}[t]{lllrrl}
\toprule
Learners & Screeners & Algorithm & Min & Max & Point estimate\\
\midrule
All & None & cSL & 0.435 & 0.539 & 0.493\\
All & None & dSL & 0.440 & 0.531 & 0.495\\
All & None & lasso & 0.498 & 0.609 & 0.550\\
All & Lasso & cSL & 0.446 & 0.564 & 0.501\\
All & Lasso & dSL & 0.430 & 0.556 & 0.500\\
All & All (-lasso) & cSL & 0.429 & 0.542 & 0.489\\
All & All (-lasso) & dSL & 0.428 & 0.542 & 0.491\\
All & All & cSL & 0.430 & 0.548 & 0.489\\
All & All & dSL & 0.430 & 0.542 & 0.491\\
All & All (+none) & cSL & 0.422 & 0.554 & 0.491\\
All & All (+none) & dSL & 0.430 & 0.542 & 0.491\\
\bottomrule
\end{tabular}}
\end{table}

In Table~\ref{tab:data_analysis_supp_all}, we display the CV R-squared (continuous outcome) and CV AUC (binary outcome) for all candidate screener-learner pairs in all super learners. The generalize linear regression model often did not converge, because there were more variables than observations. The performance of the other algorithms varied.

\begingroup\fontsize{9}{11}\selectfont

\begin{longtable}[t]{lllllll}
\caption{Estimates of cross-validated R-squared (continuous outcome), 
                                AUC and negative log likelihood (binary outcome) for the learner-screener pairs under each combination of learners and screeners. For screeners, `None' denotes no screeners; `Lasso' denotes only a lasso screener; `All (-lasso)' denotes random forest, rank-correlation, and correlation-test p-value screening; `All' denotes these three screener types plus the lasso; and `All (+none)' denotes all screeners plus the `none' screener.\label{tab:data_analysis_supp_all}}\\
\toprule
Learners & Screeners & Algorithm & Min & Max & Point estimate [95\% CI] & Risk measure\\
\midrule
\endfirsthead
\caption[]{Estimates of cross-validated prediction performance for all screener-learner pairs within the super learner. \textit{(continued)}}\\
\toprule
Learners & Screeners & Algorithm & Min & Max & Point estimate [95\% CI] & Risk measure\\
\midrule
\endhead

\endfoot
\bottomrule
\endlastfoot
All & None & SL.glm\_All & < -10 & < -10 & < -10 & R-squared\\
All & None & SL.ranger\_All & 0.38 & 0.38 & 0.380 [0.364, 0.397] & R-squared\\
All & None & earth\_1\_All & -0.258 & -0.258 & -0.258 [NaN, NaN] & R-squared\\
All & Lasso & SL.glm\_screen.glmnet & 0.158 & 0.158 & 0.158 [0.131, 0.191] & R-squared\\
All & Lasso & SL.ranger\_screen.glmnet & 0.387 & 0.387 & 0.387 [0.366, 0.409] & R-squared\\
All & Lasso & earth\_1\_screen.glmnet & -0.054 & -0.054 & -0.054 [NaN, NaN] & R-squared\\
All & All (-lasso) & SL.glm\_screen.corRank.10 & 0.192 & 0.192 & 0.192 [0.183, 0.200] & R-squared\\
All & All (-lasso) & SL.glm\_screen.corRank.25 & 0.139 & 0.139 & 0.139 [0.130, 0.149] & R-squared\\
\end{longtable}
\endgroup{}

\vspace{0.1in}

{\small
\bibliographystyle{chicago}
\bibliography{brian-papers}
}

\end{document}